\documentclass[10pt]{amsart}
\usepackage{times}
\usepackage{multicol}
\usepackage{graphics,epsf,pstricks,multicol,pst-grad,color}
\usepackage[centertags]{amsmath}
\usepackage{amsfonts}
\usepackage{amssymb}
\usepackage{amsthm}
\usepackage{newlfont}
\usepackage{ae} 
\usepackage[T1]{fontenc}
\usepackage{graphicx}
\usepackage[ansinew]{inputenc}

\newtheorem{thm}{Theorem}[section]

\newtheorem{lem}[thm]{Lemma}

\numberwithin{equation}{section}

\newcommand{\ket}[1]{\left|#1\right\rangle}
\newcommand{\bra}[1]{\left\langle#1\right|}

\newcommand{\braket}[2]{\left\langle#1|#2\right\rangle}
\newcommand{\ketbra}[2]{\ket{#1}\bra{#2}}

\newcommand{\conj}[1]{#1^{*}}
\newcommand{\abs}[1]{\left\vert#1\right\vert}

\newcommand{\bchi}{\bra{\chi}}
\newcommand{\bpsi}{\bra{\psi}}

\newcommand{\kphi}{\ket{\phi}}

\newcommand{\kchi}{\ket{\chi}}
\newcommand{\kpsi}{\ket{\psi}}

\newcommand{\kpsix}[1]{\ket{\psi(#1)}}

\newcommand{\expect}[1]{{\langle #1 \rangle}}

\newcommand{\hilb}{\mathcal{H}}
\newcommand{\compl}{\mathbb{C}}
\newcommand{\mtxt}[1]{\quad\mbox{\rm #1}\quad}
\newcommand{\mwith}{\mtxt{with}}

\newcommand{\mand}{\mtxt{and}}

\newcommand{\tr}{\mathrm{tr}\,}

\newcommand{\matr}[2]{
  \left(\begin{array}{#1}
  #2
  \end{array}\right)
}
\newcommand{\bchart}[1]{
  \begin{table}[hbt]
    \begin{center}
      \begin{tabular}{#1}
      \hline
}
\newcommand{\echart}[2]{
      \hline
      \end{tabular}
      \small\caption{\it #1}\label{#2}
    \end{center}
  \end{table}
}
\newcommand{\chart}[4]{\bchart{#1}#2\echart{#3}{#4}}

\begin{document}


\title{Conjugate Variables as a Resource
in Signal and Image Processing}%

\author{Michael N\"olle and Martin Suda \\
\\
AIT--Austrian Institute of Technology \\
Safety \& Security Department \\
Optical Quantum Technologies
}
%

\thanks{We would like to thank Jozef Gruska and Matthias Jakob for all the
valuable discussions
as well as Ian Glendinning and Janice Knight for editing the
manuscript and the discussions thereabout. We also want to thank our
company, the AIT, that enabled this research and the tax payers in
Austria who payed for it. Last but not least we thank the reviewers
who declined our project proposals on the subject and those who
fostered it (to no avail) as they prompted us to show the value in
this research even harder.}

\begin{abstract}
In this paper we develop a new technique to model joint
distributions of signals. Our technique is based on quantum
mechanical conjugate variables. We show that the transition
probability of quantum states leads to a distance function on the
signals. This distance function obeys the triangle inequality on all
quantum states and becomes a metric on pure quantum states. Treating
signals as conjugate variables allows us to create a new approach to
segment them.

\end{abstract}

\maketitle
\section{Introduction}
\pagestyle{myheadings}%
\markboth{Conjugate Variables}{Michael N\"olle, Martin Suda}%

Quantum information theory and application areas based upon it,
such as quantum computing or quantum cryptography have attracted a massive research interest and are developing rapidly. Quantum computing and quantum cryptography are primarily
concerned with the impact that the nature of physical quantum systems has
on the respective fields of computing and cryptography. Using physical systems, one
has to obey the restrictions imposed by nature on physical
quantum states, for example that they are not directly observable. They can be measured, but in general a measurement only reveals parts of the
information contained in the quantum state. 

Some authors have applied quantum information theory to statistics and probability theory (\cite{Barn2003,BroHug2006}) as well as signal and image processing applications
(\cite{Cof2006,EldOpp02}). Their work focuses mainly on the abstract
mathematical concept of quantum information theory and might be
referred to as Quantum Information processing Algorithms (QIA) in analogy to
the term Quantum Signal Processing created by Eldar (\cite{EldOpp02}). 
Applying the formalism of quantum information theory to information processing on classical computers can result in novel but still 'classical' algorithms. 

In this paper we follow the QIA approach, motivated by the fact that we have found considerably
easier descriptions of well suited algorithmic solutions for the problems at hand in several application areas (\cite{Nol05,Nol07a,Nol07b}) within the mathematical setting of the
quantum space. One reason might be that we were forced to think more deeply about the normalization of data and observations and the metrical functions to be used. A crucial factor seems to be the ability to represent some relations between
two (or more) sources of information jointly as quantum information, i.e. in a
quantum state. In analogy to the behaviour of physical
conjugate variables, e.g. position and momentum, which play a
fundamental role in quantum mechanics, we term this a
$\textsl{conjugate information variable}$. In essence we refer to
two sources of information as being conjugate to each other if they are
in a special relation: Whenever one of them becomes highly
predictable the other one is either undefined or unpredictable and
vice versa.

The special relation between physical observations of conjugate
quantities was an important factor in the development of
quantum mechanics in the early $20^{th}$ century. In the quantum
mechanical setting, Heisenberg's uncertainty principle expresses this
relation, e.g. between momentum and position. To be more precise,
momentum and position are represented as operators in quantum mechanics and they are related to each other by derivatives. Here an exact definition of
uncertainty means that these operators do not commute.

Our main justification for using the same mathematical structure, i.e.
Hilbert spaces, is that information sources exist, which have a similar conjugate relation. As an illustrative example consider we want to
observe two species of birds lets say one is blue the other one
green and their chirp is quite distinct but cannot be reliably heard
during day time due to background noises which are absent at
night. Obviously during the day we use our eyes and at night our
ears to distinguish the species. But is there a seamless way to fuse
both signals that additionally helps us to distinguish them at
sunrise or sunset with higher reliability? 

Or more seriously,
consider a simple classification task where we want to decide whether
parts of a function are locally constant, rising,
falling or minimum or maximum. We might want to use the derivatives
of the function to achieve this, as we know
that when the first derivative becomes zero the entire information on the
extreme points of the function relies on the sign of the second (or
third...) derivative. On the other hand when the second derivative
becomes zero, the information whether we found a point of (rising or
falling) inflection or the function is constant at this point
depends entirely on the first (or third...) derivative. Again it
would be desirable to represent the first and second derivative such
that in 'grey areas' in between the two extremes
either one alone (if it is more reliable) or both may contribute to the classification result.

The authors are fully aware that most of the material on theoretical
quantum information is already covered elsewhere. In particular we
refer to the textbooks of Gruska \cite{Gru99} and Nielsen
and Chuang \cite{Niel00}. Nonetheless we find it worthwhile to
summarize some of the basic ideas behind quantum information theory
in order to help readers who are not too familiar with them and the
notation used in this area. We will omit reoccurring references to
\cite{Gru99,Niel00}, where most of the material is treated
in-depth, and indicate when the result can be found elsewhere. This
introduction to quantum information theory is contained in Chapter
\ref{ChapQIT} whereas Chapter \ref{ChapQIA} analyzes the encoding of
information into quantum states. In Chapter \ref{ChapQBIT} we take a
closer look on two-dimensional systems. A new approach to signal
segmentation is given in Chapter \ref{ChapApproach}. We conclude the
paper with some discussion and references to future work (Chapters
\ref{ChapDis}, \ref{ChapFut}).

The fact that in most parts of this paper we consider
random variables and distributions rather than signals reflects the fact that we assume the signals to be normalized. In quantum states/signals the same assumption is usually made.

\section{An Introduction to Quantum Information Theory}\label{ChapQIT}

\subsection{Quantum Systems, Quantum States and Qubits}\label{SecQSpace}

The first postulate of quantum mechanics states that any physical
system can be described by a unit vector in an associated complex
vector space which is called Hilbert space $\hilb$. A given unit
vector $\vec{\psi}$ is called the state of the system. In this paper
we restrict ourselves to finite dimensional Hilbert spaces. The
simplest non-trivial system is a two-dimensional system with the
state space $\hilb=\compl^2$. Such systems are called {\it quantum
bits} or {\it qubits}. In the usual mathematical notation, the state
of a qubit can thus be written as the vector
\begin{equation}
\vec{\psi}=\matr{c}{\alpha \\ \beta}=\alpha\, \hat{e}_0 + \beta\,
\hat{e}_1 \mwith \alpha,\beta\in\compl \mand |\alpha|^2+|\beta|^2=1,
\end{equation}
where the basis vectors $\hat{e}_0$ and $\hat{e}_1$ are orthonormal
unit vectors. In quantum computing, it is customary to use the Dirac
(or ``bra-ket'') notation: The column ``ket''-vectors
\begin{equation}
\ket{0}\equiv\hat{e}_0=\matr{c}{1\\0}
\mand\ket{1}\equiv\hat{e}_1=\matr{c}{0\\1},
\end{equation}
or generally $\ket{i}\equiv\hat{e}_i$ for higher dimensional cases,
form the {\it computational} or{ canonical basis} of $\hilb$ and the
above state can be written as a {\it superposition}
\begin{equation}\label{psi}
\kpsi=\alpha\,\ket{0}+\beta\, \ket{1}.
\end{equation}

The {\it dual} ``bra''-vector $\bpsi$ corresponds to the associated
row-vector $(\conj\alpha,\conj\beta)$ with complex conjugate
components, thus
\begin{equation}
\bpsi=\conj\alpha\,\bra{0}+\conj\beta\, \bra{1}.
\end{equation}

The scalar product $\braket{\psi}{\chi}$ and the outer product
$\kpsi\bchi$ are thus reduced to matrix multiplications.
Any linear operator $\mathbf{A}: \hilb \rightarrow \hilb$ over
$\hilb=\compl^n$ can be written as
\begin{equation}\label{DefOperator}
    \mathbf{A} = \sum_{i,j}^{n-1} a_{i,j} \ket{i}\bra{j}.
\end{equation}
For a {\it Hermitian operator} $\mathbf{A} = \mathbf{A}^\dag$ holds,
where in matrix notation $\mathbf{A}^\dag$ is the transpose of
$\mathbf{A}$ with conjugate complex entries. Whenever we omit the
index bounds, the sum is from $0,\ldots, n-1$, where $n$ is the
dimension of the considered Hilbert space.

The orthonormality and completeness of the computational basis can
be written as
\begin{equation}
\braket{i}{j}=\delta_{ij}\mand \sum_i \ket{i}\bra{i}=\mathbb{I},
\end{equation}
where $\delta_{ij}$ denotes the Kronecker delta function and
$\mathbb{I}$ is the identity. Table \ref{dirac} summarizes
frequently-used Dirac notation and its meaning. 
\chart{|c|l|}{
  {\bf Notation} & {\bf Description} \\\hline
  & \\
  $\kpsi$ & general ``ket'' vector, e.g. $\kpsi=( c_0, c_1, \ldots )^T$ \\
  $\bpsi$ & dual ``bra'' vector to $\kpsi$, e.g. $\bpsi=( c_0^*, c_1^*, \ldots )$ \\
  $\ket{n}$ & $n^\mathrm{th}$ basis vector of computational basis $N=\{\ket{0},\ket{1},\ldots\}$ \\
  $\braket{\phi}{\psi}$ & inner product of $\kphi$ and $\kpsi$ (scalar) \\
  $\kphi\bpsi$ & outer product of $\kphi$ and $\kpsi$ (matrix) \\
  $\kphi\kpsi$ & tensor product $\kphi\otimes\kpsi$ (vector) \\
  $\ket{i,j}$ & tensor product $\ket{i}\ket{j}$ of the basis vectors $\ket{i}$ and $\ket{j}$ \\\hline
}{Dirac (``bra-ket'') Notation}{dirac}

\subsection{Measurement, von Neumann Measurement}\label{Measurement}

From a physical point of view any information contained in a quantum
state $\kpsi$ can only be accessed via a $measurement$ which is
described by a set of (hermitian) operators adding up to the
identity operator (completeness condition):
\begin{equation}\label{Completeness}
\sum_{i=0}^{m-1} \mathbf{M}_i = \mathbb{I}.
\end{equation}
As a result of a measurement on $\kpsi$ we get an index
$i\in\{0,\ldots,m-1\}=S$ with some probability. A standard
interpretation of a measurement is that the state $\kpsi$
`collapses' onto the post-measurement state
$\frac{1}{\sqrt{p(i)}}M_i\kpsi$ with probability $p(i)=\bpsi
M_i\kpsi$. It should be noted that the idea that the quantum state
spontaneously is changed via a measurement has been the subject of much debate ever since it was proposed in the 1950s. We do not want to address this issue,
as the material discussed here is unaffected by it. The
fact that the only direct information we can gain from a physical quantum system is the outcome $i$ of the measurement is undisputed. Reapplying the measurement does not change this result any further. This means in particular that we cannot gain any direct information about the co-ordinates of a quantum state.

However if we are given a large number of 'identical' quantum states and we measure them one by one, we eventually gain the frequency of each index and after proper normalization the probability information $p(i)$ for all $i\in S$.
Taking the completeness condition (\ref{Completeness}) into account
we see that the probabilities properly add up to one:
$\bpsi\left(\sum_i^m \mathbf{M}_i\right)\kpsi$ $= \bpsi
\mathbb{I}\kpsi = 1$. In slightly different terms we get the
probability distribution of a discrete random variable over the
given index set $P(X=i)=p(i)$. The values the random variable
can take\footnote{the elements of the $\sigma$ algebra.} and the
index of the measurement operators are in a one to one
correspondence. The resulting probability distribution is fixed as
soon as we are given the state $\kpsi$ and the set of measurement
operators $\mathbf{M} =\{\mathbf{M}_{i=0}^{m-1}\}$. In the next
section we will see that the converse in general is not true. That
is, a given probability distribution does not uniquely determine a
quantum state and/or a set of measurement operators.

In this paper we will restrict ourselves to a particulary simple form of measurement which is described by projectors, the so-called von Neumann measurement. Given a unit vector $\kpsi$ the projector $\textbf{P}$ onto $\kpsi$ takes the form
\begin{equation}\label{Projector}
    \textbf{P} = \kpsi\bpsi.
\end{equation}
Applying $\textbf{P}\kpsi=\kpsi\braket{\psi}{\psi}=1\cdot\kpsi$ we
see that it projects $\kpsi$ onto itself and any other
unit vector $\kchi$ is projected onto:
$\kpsi\braket{\psi}{\chi}=\braket{\psi}{\chi}\kpsi$, i.e. in the
direction of $\kpsi$ times the scalar $\braket{\psi}{\chi}$. 
If $\textbf{P}^\bot = \mathbb{I}-\textbf{P}$ is the
projector onto the orthogonal complement space of $\textbf{P}$,
$\{\textbf{P},\textbf{P}^\bot\}$ forms a measurement. For any state
$\kchi$ the probability that it is changed into $\kpsi$ via this
measurement is
\begin{equation}\label{Projection}
    p\left(\kchi \rightarrow\kpsi\right) \equiv \bchi\textbf{P}\kchi = \braket{\chi}{\psi}\braket{\psi}{\chi} =
    \left| \braket{\psi}{\chi} \right|^2,
\end{equation}
whereas
\begin{equation}\label{ComplementProjection}
    \bchi\textbf{P}^\bot\kchi =
    \bchi\left(\mathbb{I}-\textbf{P}\right)\kchi = 1-\left| \braket{\psi}{\chi} \right|^2
\end{equation}
gives the probability that the post measurement state lies in the
orthogonal complement space. $p\left(\kchi
\rightarrow\kpsi\right)=\left| \braket{\psi}{\chi} \right|^2$ is
called the transition probability. In the next section we will
analyze these equations further.

\subsection{Operator Space, Density Operator}
The quantum state space under consideration is usually called the
principal system. The $\textsl{operator space}$ over a principal
system is given by all linear mappings of this state space onto
itself, i.e. given the state space $\hilb$ the operator space is
$\hilb\otimes\hilb=L_\hilb$. The inner product in the operator space
takes the form of the Hilbert-Schmidt inner product:
\begin{equation}\label{HilbertSchmidtInnerProduct}
    \left(\textbf{A};\textbf{B}\right) =
    \tr\left(\textbf{A}^\dag\textbf{B}\right);\quad
    \textbf{A}, \textbf{B}\in\hilb\otimes\hilb,
\end{equation}
where the trace of an operator is the sum of the diagonal elements
of the operator written as a matrix. In terms of the
Dirac notation the trace of an operator may be written as
\begin{eqnarray}
\nonumber \tr\left(\textbf{A}\right) &=& \tr\left(\sum_{i,j} a_{i,j}
\ket{i}\bra{j}\right)
=\sum_{k}\bra{k}\left(\sum_{i,j} a_{i,j}
\ket{i}\bra{j}\right)\ket{k}
\\
&=& \sum_{k}^n a_{k,k}. \label{Trace}
\end{eqnarray}
Equation \ref{HilbertSchmidtInnerProduct} enables us to define a
$\textsl{norm}$ on $L_\hilb$:
  \begin{eqnarray}
\nonumber
    \|\textbf{A}\|^2 &=& tr(\textbf{A}^\dag \textbf{A})
          = \tr\left(\left(\sum_{i,j}a_{i,j}\ket{i}\bra{j}\right)^\dag
                             \sum_{l,m}a_{l,m}\ket{l}\bra{m}\right)
                             \\
\nonumber
           &=& \tr\left(\sum_{i,j}a^*_{i,j}\ket{j}\bra{i}
                       \sum_{l,m}a_{l,m}  \ket{l}\bra{m}\right)
                             \\
\nonumber
           &=& \tr\left(\sum_{i,j,m}a^*_{i,j}a_{i,m}\ket{j}\bra{m}\right)
           =\sum_{i,j}a^*_{i,j}a_{i,j}.
                             \\
    \|\textbf{A}\| &=& \sqrt{\sum_{i,j}a^*_{i,j}a_{i,j}} = \sqrt{\sum_{i,j}|a_{i,j}|^2}.
           \label{EuklidianNorm}
  \end{eqnarray}
(\ref{EuklidianNorm}) is called the {\it Frobenius} norm or {\it
Hilbert-Schmidt} norm of $L_\hilb$. Consequently, for operators
$\textbf{A}, \textbf{B}\in L_\hilb$ we may calculate the
$\textsl{Euclidian distance}$ $\|\textbf{A}-\textbf{B}\|$ as:
\begin{eqnarray}
\nonumber
 \|\textbf{A}-\textbf{B}\| &=&
    \sqrt{\sum_{i,j}(a^*_{i,j}-b^*_{i,j})(a_{i,j}-b_{i,j})}
  = \sqrt{\sum_{i,j}\left[|a_{i,j}|^2+|b_{i,j}|^2-2\Re\left(a_{i,j}b^*_{i,j}\right)\right]} \\
                          &=&
                          \sqrt{\left[\|\textbf{A}\|^2+\|\textbf{B}\|^2-
                          2\Re\tr\left(\textbf{A}\textbf{B}^\dag\right)\right]},
                          \label{EuklidianDistance}
\end{eqnarray}
where $\Re$ gives the real part of a complex number.

Finally we would like to emphasis the well known fact that global
phase changes of a state do not affect the expectation value of an
operator.

\begin{lem}[Global phase invariance of the expectation of linear operators]
Let $A\in L_\hilb$ be a linear operator. Then the expectation of $A$
is invariant with respect to global phase changes, i.e. for any
$\tilde{\ket{\psi}}= e^{i\varphi}\ket{\psi}$,
$-\pi\le\varphi\le\pi$, $\ket{\psi}\in\mathbb{C}^n$:
\begin{equation*}
    \tilde{\bra{\psi}}\textbf{A}\tilde{\ket{\psi}}=\bra{\psi}\textbf{A}\ket{\psi}.
\end{equation*}
\end{lem}
This is easily verified by
\begin{proof}
$\tilde{\bra{\psi}}\textbf{A}\tilde{\ket{\psi}}=
e^{-i\varphi}\bra{\psi}\textbf{A} e^{i\varphi}\ket{\psi}=
e^{-i\varphi+i\varphi}\bra{\psi}\textbf{A}\ket{\psi}=
\bra{\psi}\textbf{A}\ket{\psi}.$
\end{proof}

After these general remarks on the operator space we will turn our
attention to a special subset of operators, the so called
$\textsl{density operators}$. Lets first consider the single qubit
state
\begin{equation}
  \ket{+}=\frac{1}{\sqrt 2}\ket{0}+\frac{1}{\sqrt 2}\ket{1}.
\end{equation}

When we measure $\ket{+}$ by projectors onto the computational basis
$\{\ket{0}\bra{0},\ket{1}\bra{1}\}$, we will find that either of the
two possible post-measurement states $\ket{0}$ and $\ket{1}$ are
equally likely with a transition probability of
$p_0=p_1=\frac{1}{2}$. So the experiment is equivalent to a
classical flip of an unbiased coin and we might be tempted to think
that $\ket{+}$ in fact describes a state where the system state is
{\it unknown} which is either $\ket{0}$ or $\ket{1}$ with equal
probability.

If, however, we measure via $\{\ket{+}\bra{+},\ket{-}\bra{-}\}$ with
\begin{equation}
 \ket{-}=\frac{1}{\sqrt 2}\ket{0}-\frac{1}{\sqrt 2}\ket{1},
\end{equation}
on the same state $\ket{+}$, then the outcome is deterministic as
$\ket{+}$ is an eigenvector of $\ket{+}\bra{+}$. Obviously, the
non-determinism of a quantum measurement is inherently different from
the classical randomness of (un)biased coin flips.

There exists an elegant and consistent way to capture both concepts
of randomness: Assuming we know that a quantum system is in one of
$k$ states $\ket{\psi_i}$ with the probabilities $p_i$, then we can
aggregate our knowledge of the state in a (semi) positive {\it
density operator}
\begin{equation}
  \rho=\sum_{i=1}^k p(i) \ket{\psi_i}\bra{\psi_i}
  \mwith \sum_{i=1}^k p(i)=1.
\end{equation}

As the probabilities add up to 1 and by the linearity of the trace,
it is assured that $\tr\rho=\tr\left(\sum_{i=1}^k p_i
\ket{\psi_i}\bra{\psi_i}\right)=\sum_{i=1}^k p_i
\tr\ket{\psi_i}\bra{\psi_i}=\sum_{i=1}^k p_i=1$ where we used Eq. \ref{Trace}. If $\tr\rho^2=1$
then $\rho$ can be written in the form $\rho=\kpsi\bpsi$ and the
system is said to be in a {\it pure state}. If $\tr\rho^2<1$, then
$\rho$ is referred to a {\it mixed state}. The transition
probability $p(\rho \rightarrow\chi)$ to reduce $\rho$ (pure or mixed state) to the pure state 
$\rho'=\kchi\bchi$ when performing a measurement in a basis
containing $\kchi$ is the {\it expectation value}
\begin{equation}\label{transprob2}
  p(\rho \rightarrow\chi)=\expect{\rho}_\chi=\bchi\rho\kchi.
\end{equation}
Note that for pure states, (\ref{transprob2}) is equivalent to
(\ref{Projection}), as we would expect.

There are some properties of density operators that should be
mentioned:
\begin{itemize}
  \item $\tr\rho = 1$ (see above);
  \item $\rho=\rho^\dag$, i.e. density operators are hermitian
  operators;
  \item any density operator has a spectral decomposition, $\rho =
  \textbf{U}\textbf{D}\textbf{U}^\dag$, where $\textbf{U}$ is
  a unitary matrix, i.e. $\textbf{U}^\dag\textbf{U} = \mathbb{I}$, and $\textbf{D}$ is a diagonal matrix with
  nonnegative real diagonal elements. Therefore, $\rho = \sum_i d(i)
  \ket{u_i}\bra{u_i}$, with $\sum_i d(i)=1$, where $\ket{u_i}$ is the basis generated by the columns of $U$
  and $d_i$ are the diagonal entries of $D$. The $\ket{u_i}$ are called
  the eigenbasis of $\rho$;
  \item consider two density operators in their eigenbasis representation: \\
  $\rho = \sum_i p(i)\ket{\psi_i}\bra{\psi_i}$ with $\sum_i p(i)=1$ and
$\sigma = \sum_i q(i)\ket{\chi_i}\bra{\chi_i}$ with $\sum_i q(i)=1$,
then $\tr\left(\rho\sigma\right) = \sum_{i,j} p(i)q(j)
  \left|\braket{\psi_i}{\chi_j}\right|^2.$
  This implies, as $p_i$, $q_j$ and $\left|\braket{\psi_i}{\chi_j}\right|^2$ are nonnegative and real numbers, that
  $\tr\left(\rho\sigma\right)=\Re\tr\left(\rho\sigma\right)$;
  \item $0\le \tr\left(\rho\sigma\right) \le 1$. The lower bound
  follows again from the fact that the sum is over nonnegative and
  real numbers. To verify the upper bound we substitute $\left|\braket{\psi_i}{\chi_j}\right|^2\le
  1$, which gives
  \begin{equation}\label{DensityOperatorTraceUpperBound}
    0\le \tr\left(\rho\sigma\right) = \sum_{i,j} p(i)q(j) \left|\braket{\psi_i}{\chi_j}\right|^2
    \le \sum_{i,j} p(i)q(j) = 1.
  \end{equation}

\end{itemize}
The minimum error discrimination of mixed quantum states is an
interesting and well studied problem. We refer the interested reader
to pioneering work of \cite{Hel76,Hol82} and further results in
\cite{Fuch96,Ber04} and the literature cited therein.

%
%
%

\section{Encoding classical Information as Quantum
Information}\label{ChapQIA}

In Section \ref{Measurement} we have seen that given a set of
measurement operators $\mathbf{M} =\{\mathbf{M}_{i=0}^{m-1}\}$ and a
quantum system $\kpsi$ uniquely determines a probability
distribution $P(X=i)=p(i), i=0,\ldots, m-1$. In this section we want
to analyze the converse, i.e. given a probability distribution we
want to determine quantum states and measurement operators that
produce it.

Let us consider a random variable $X$. The set of values $X$
can take is given by $S$ and for any probability distribution
\begin{equation}\label{EqNormalisationCondition}
    \sum_{x\in S} p(X=x) = 1,\quad p(X=x)\geq 0, \mtxt{for all $x\in S$.}
\end{equation}
A given probability distribution may be interpreted as vector from
the $n=\left| S \right|$ dimensional real vector space
$\mathbb{R}^{n}$. In this case the basis of the space can be indexed
by the elements of $S$ and $\ket{x}, x\in S$ denotes a basis
vector. In this basis we define
$\overrightarrow{p(X)} \equiv \ket{p(X)} = \sum_{x\in S} p(x)
\ket{x}$. Due to the linearity of the normalization condition
(\ref{EqNormalisationCondition}) the set of all probability
distributions forms a $n-1$ dimensional hyper-plane, sometimes
called simplex, in this space.

As we would like to make a connection between the set of all
probability distributions in $\mathbb{R}^{n}$ and quantum states
from a complex Hilbert space $\mathbb{C}^{n}=\mathcal{H}$, we
formerly introduce complex numbers $\alpha_{\varphi_{x}}\in
\mathbb{C}$ such that
\begin{equation}\label{EqManifoldMapping}
p(x) = \conj{\alpha_{\varphi_{x}}}\alpha_{\varphi_{x}} =
|\alpha_{\varphi_{x}}|^2.
\end{equation}
The choice of $\alpha_{\varphi_{x}}$ is not unique. Using Euler's
Equation on
\begin{equation*}\label{EulerEquation}
\alpha_{\varphi_{x}} = \sqrt{p(x)}e^{i\varphi_{x}} =
\sqrt{p(x)}(\cos\varphi_{x} + i\sin\varphi_{x}),
-\pi\le\varphi_{x}\le\pi,
\end{equation*}
Equation (\ref{EqManifoldMapping}) can be rewritten as
\begin{eqnarray}
p(x) = \conj{\alpha_{\varphi_{x}}}\alpha_{\varphi_{x}} &=&
p(x)\cos^2\varphi_{x} + p(x) \sin^2\varphi_{x}.
\label{EqManifoldEuler}
\end{eqnarray}

It is noteworthy that all numbers
$\alpha_{\varphi_{x}}$ fulfilling (\ref{EqManifoldMapping}) lie on a
complex sphere with radius $\sqrt{p(x)}$ as can be verified from
(\ref{EqManifoldEuler}).

Already here we see that encoding probabilities as quantum states
requires some more information than only the probability
distribution if we want to be able to adress all quantum 
states\footnote{To be more precise: we need to know the
meaning of '$-\sqrt{p(x)}$'. This cannot be derived from the
probability distribution alone. One bit of additional information is required.
If this bit of information is available, the complex part insures a
smooth transition from $\sqrt{p(x)}$ to $-\sqrt{p(x)}$, i.e. without
changing the probability of $\ket{x}$ if measured in the same
basis.}. If we restrict the encoding onto the real nonnegative
square root we gain the same quantum state as described in
\cite{Gro02}.

Consider the (state) vector $\ket{\psi} \in \mathbb{C}^{n}$:
\begin{equation}\label{EqStateVector}
    \ket{\psi}=\sum_{x\in S}\alpha_{\varphi_{x}}\ket{x}.
\end{equation}
The set of all projectors onto basis states, $\{\textbf{P}_{\ket{x}}
= \ket{x}\bra{x},x\in S\}$, taken as measurement operators result in
the probability distribution $p(X)$ when $\ket{\psi}$ is measured,
and fulfill the completeness condition (\ref{Completeness}).

Using (\ref{EqManifoldEuler}) and the normalization condition
(\ref{EqNormalisationCondition}) we see that $\ket{\psi}$ is a unit
vector in $\mathbb{C}^{n}$.
These unit vectors are related to probability distributions via (\ref{EqManifoldMapping})
and lie on a complex unit hyper-sphere centered at the origin.

We will call the assignment
\begin{equation}\label{QuantumEncoding}
    \emph{\textsl{E}}\left(p\left(X=x\right),C\right) \rightarrow \alpha_{\varphi_{x}}
\end{equation}
$\textsl{quantum encoding}$ of $p(X)$ with respect to the
\textsl{conjugate information} $C$. If we want to decode the
probability information of the quantum encoding we have to apply
$\{\textbf{P}_{\ket{x}} = \ket{x}\bra{x},x\in S\}$ to $\ket{\psi}$.

\subsection{Distances}
When given two probability distributions over the same index set a
natural and important question is: how similar or how close are
they? The answers to this question can be quite diverse depending on
the scientific discipline they have been derived from. Table
\ref{ProbabilityDistance} gives an overview on the more frequently
used probability distribution distance measures as they can be found
in the literature.

\chart{|c|l|c|}{
  {\bf Name} & {\bf Formula} & {\bf Metric} \\\hline
  Euclidian distance & $\sqrt{\sum_i (p(i)-q(i))^2}$ & yes \\\hline
  trace or & $\sum_i |p(i)-q(i)|$ & yes \\
  variational distance (\cite{Lin91})&  & \\\hline
  relative $\chi^2$ & $\sum_i \frac{(p(i)-q(i))^2}{2(p(i)+q(i))}$ & $\chi$ yes \\\hline
  relative entropy or Kullback-& $\sum_i p(i)\log\frac{p(i)}{q(i)} =$ & no \\
  Leibler divergence (\cite{Kul97}) &  $-H(p)-\sum_i p(i)\log q(i)$ & \\\hline
  Jensen-Shannon divergence or & $JS(p,q)=$
  & $\sqrt{JS(p,q)}$ \\
   symmetric relative entropy (\cite{Lin91}) 
& $H\left(\frac{1}{2}(p+q)\right)-\frac{1}{2}\left(H(p)+H(q)\right) $ & yes \\\hline
  Bhattacharyya distance (\cite{Aher97,Kul97})& $1-\sum_i \sqrt{p(i)}\sqrt{q(i)}$  & yes\\
   \hline
}{Frequently used probability distribution distance measures. $p$
and $q$ are probability distributions over the same index set. $H(p)
= -\sum_i p(i) \log p(i)$ is the Shannon entropy
(\cite{Cov91,Gal68,Ash65}).}{ProbabilityDistance}

Likewise we find adapted distance measures for quantum states and/or
density operators (\cite{Fuch96}). Based on the transition
probability of quantum states we may define a distance function:

\begin{lem}
\label{LemmaNoNameMetric}
Let $\rho, \sigma \in L_\hilb$ be density operators,
($\tr\rho=\tr\sigma=1$). Then the function
\begin{equation*}
    D(\rho,\sigma) = \sqrt{1-\tr\left(\rho\sigma\right)}
\end{equation*}
satisfies the following properties:
\begin{enumerate}
  \item $0\le D(\rho,\sigma)\le 1$,
  \item $D(\rho,\sigma) = D(\sigma,\rho)$, (symmetry)
  \item for any density operator $\tau\in L_\hilb, \tr\tau = 1$:
  \begin{equation*}
    D(\rho,\sigma) \le D(\rho,\tau)+D(\tau,\sigma),\quad\textrm{(triangle inequality)}
  \end{equation*}
  \item on pure states $D(.,.)$ defines a metric.
\end{enumerate}
\end{lem}

The fact that $D$ defines a metric on pure states is well known
(\cite{Niel00}, p. 500). In the literature it is sometimes called
{\sl no-name metric} and it is closely related to the so called
Fubini study metric. Nevertheless, we would like to prove Lemma
\ref{LemmaNoNameMetric} as the proof offers some insight into the
relation of $D$ and the Euclidian metric of the operator space and
the principal system.
\begin{proof}
Property (1) of Lemma \ref{LemmaNoNameMetric} is a direct consequence of Equation
(\ref{DensityOperatorTraceUpperBound}). Property (2) follows from the
fact that the trace of an operator product is invariant against
cyclic permutations of the arguments, i.e.
$\tr(\rho\sigma)=\tr(\sigma\rho)$.

Let $\rho, \sigma, \tau \in L_\hilb$ be density operators, i.e.
$\tr\rho = \tr\sigma = \tr\tau = 1$ and $\|\rho\| \le 1, \|\sigma\|
\le 1, \|\tau\| \le 1$. To see property (3) we first note that the
Euclidian distance (see Equation \ref{HilbertSchmidtInnerProduct}) 
of the operator space fulfills the triangle
inequality for all operators in $L_\hilb$, i.e. $\|\rho-\sigma\| \le
\|\rho-\tau\| + \|\tau-\sigma\|$ holds. We calculate
\begin{eqnarray}
\nonumber
  \|\rho-\sigma\|^2 &\le& \left(\|\rho-\tau\| + \|\tau-\sigma\|\right)^2 \\
\nonumber
  \|\rho\|^2+\|\sigma\|^2 - 2\tr\left(\rho\sigma\right) &\le&
  \|\rho-\tau\|^2 + \|\tau-\sigma\|^2 + 2\|\rho-\tau\|\|\tau-\sigma\|
  \\
\nonumber
  &=& \|\rho\|^2+\|\tau\|^2 - 2\tr\left(\rho\tau\right)+
      \|\tau\|^2+\|\sigma\|^2 - 2\tr\left(\tau\sigma\right) +\\
\nonumber
   &+&  2\|\rho-\tau\|\|\tau-\sigma\|,
\end{eqnarray}
where we used Equation \ref{EuklidianDistance} in the last step.
Simplifying and expanding the mixed term gives:
\begin{eqnarray}
\nonumber
  - \tr\left(\rho\sigma\right) &\le&
      \|\tau\|^2 - \tr\left(\rho\tau\right)
      - \tr\left(\tau\sigma\right) + \\
\nonumber
 &+&
 \sqrt{\left[\|\rho\|^2+\|\tau\|^2-2\tr\left(\rho\tau\right)\right]
              \left[\|\tau\|^2+\|\sigma\|^2-2\tr\left(\tau\sigma\right)\right]}.
\end{eqnarray}
By substituting $\|\rho\|^2+\|\tau\|^2\le 2$ and
$\|\tau\|^2+\|\sigma\|^2\le 2$ we get
\begin{eqnarray}
\label{EuklidCond}
  - \tr\left(\rho\sigma\right) &\le&
      \|\tau\|^2 - \tr\left(\rho\tau\right) - \tr\left(\tau\sigma\right) + \\
\nonumber
 &+&    2\sqrt{\left(1-\tr\left(\rho\tau\right)\right)
               \left(1-\tr\left(\tau\sigma\right)\right)}.
\end{eqnarray}
With this we are ready to prove property (3) of the above Lemma. We
substitute $\|\tau\|^2\le 1$ in (\ref{EuklidCond}) and add $1$ to
both sides. Then
\begin{eqnarray}
\nonumber
 D(\rho,\sigma)^2 = 1 - \tr\left(\rho\sigma\right) &\le&
      2 - \tr\left(\rho\tau\right) - \tr\left(\tau\sigma\right) + \\
\nonumber
 &+&    2\sqrt{\left(1-\tr\left(\rho\tau\right)\right)
               \left(1-\tr\left(\tau\sigma\right)\right)} \\
\nonumber
   &=& \left(\sqrt{1-\tr\left(\rho\tau\right)}+\sqrt{1-\tr\left(\tau\sigma\right)}\right)^2 \\
\nonumber
   &=& \left( D(\rho,\tau) +D(\tau,\sigma)\right)^2.
\end{eqnarray}
Reading top to bottom we see that the triangle inequality holds for
$D(.,.)$ and all $\rho, \sigma, \tau \in L_\hilb$:
\begin{equation}\label{NoNameTriangleInEq}
    D(\rho,\sigma) \le D(\rho,\tau) +D(\tau,\sigma).
\end{equation}
For property (4) we have to verify that $D(\rho,\rho)=0$. This is only
true for pure states as in general $\tr\rho^2 \le 1$ with equality
if and only if $\rho$ is pure. On the other hand for pure states
$D(.,.)$ fulfills all the requirements of a metric.
\end{proof}

Indeed, for pure states $D$ is identical to
the Euclidian norm of the operator space up to a constant factor, which is easy to verify by
noting that for any pure state $\kpsi$ and any density operator
$\sigma$ the following holds (see. Eq. \ref{Trace}):
\begin{equation}\label{EqTracePurestateDens}
    tr\left(\ket{\psi}\bra{\psi}\sigma\right)= \bpsi\sigma\kpsi .
\end{equation}
If $\sigma$ is pure as well, i.e $\sigma = \ket{\xi}\bra{\xi}$, we
get
$tr\left(\ket{\psi}\bra{\psi}\sigma\right)=tr\left(\ket{\psi}\braket{\psi}{\xi}\bra{\xi}\right)=
\abs{\braket{\psi}{\xi}}^2$. With this and equation
\ref{EuklidianDistance} the distance computes to
$\|\ket{\psi}\bra{\psi}-\ket{\xi}\bra{\xi}\| =
\sqrt{2}\sqrt{1-\abs{\braket{\psi}{\xi}}^2}$ and
$D\left(\ket{\psi}\bra{\psi},\ket{\xi}\bra{\xi}\right)=\sqrt{1-\abs{\braket{\psi}{\xi}}^2}$.
For convenience we will write $D(\ket{\psi},\ket{\xi})\equiv
D\left(\ket{\psi}\bra{\psi},\ket{\xi}\bra{\xi}\right)$ for short
when dealing with pure states only.

On pure states the no-name metric takes therefore the form
\begin{equation}\label{EqFubini}
    D(\ket{\psi},\ket{\xi}) = \sqrt{1-|\braket{\psi}{\xi}|^2},\mwith
    \ket{\psi}=\sum_x \alpha_{\varphi_{x}}\ket{x},
    \ket{\xi}=\sum_x \beta_{\chi_{x}}\ket{x}.
\end{equation}
One may derive an intuitive geometric argument to compare two state
vectors in $\mathbb{C}^{n}$ (see Section \ref{PhaseDropping} and
Figure \ref{FigExample4}). The vectors $\ket{\psi}$ and $\ket{\xi}$
in general span a two-dimensional complex plane and the intersection
of this plane with the hyper-sphere generates a complex unit circle.
If $\ket{\psi}=\ket{\xi}$ the intersection degenerates into two
points.

{\sl Remark (a word of care):} Given two probability distributions
of the same random variable $S$ way may construct corresponding
(diagonal) density operators by: $\rho=\sum_{x\in
S}p(x)\ket{x}\bra{x}$, $\sigma=\sum_{x\in S}q(x)\ket{x}\bra{x}$.
Their Euclidian operator distance $\|\rho-\sigma\|^2=
tr(\rho^2)+tr(\sigma^2)- 2 tr(\rho\sigma)$ is equal to the Euclidian
distance of the probability distributions. But $\|\rho-\sigma\|^2
\neq 2(1-tr(\rho\sigma)) = 2 D(\rho,\sigma)^2$ unless both
distributions are pure, i.e. for both probability distributions
there exists one event, lets say $x_p,x_q$, such that
$p(x_p)=q(x_q)=1$.

For the rest of the chapter we will assume pure states only.

\subsubsection{Projection, transition probability}
Let $\ket{\psi}=\sum_{x\in S}\alpha_{\varphi_{x}}\ket{x},
\alpha_{\varphi_{x}} = \sqrt{p(x)} e^{i\varphi_{x}}$ and
$\ket{\xi}=\sum_{x\in S}\beta_{\chi_{x}}\ket{x}, \beta_{\chi_{x}} =
\sqrt{q(x)} e^{i\chi_{x}}$. As we have seen in Section
\ref{Measurement}, Equation (\ref{Projector}) the projector
$\textbf{P}$ onto $\ket{\psi}$ is given by $\textbf{P} =
\ket{\psi}\bra{\psi}$ and the transition probability
\begin{equation*}
    p\left(\ket{\xi}\rightarrow\ket{\psi}\right) = \bra{\xi}\textbf{P}\ket{\xi} =
    \braket{\xi}{\psi}\braket{\psi}{\xi} =
    |\braket{\psi}{\xi}|^2
\end{equation*}
gives the square of the length of the projection of $\ket{\xi}$ onto
$\ket{\psi}$. Substituting $\gamma_x =\varphi_{x}-\chi_{x}$ and
using Euler's Equation, we get:
\begin{eqnarray}
\nonumber
 \braket{\xi}{\psi} &=&
    \sum_{x\in S}\sqrt{p(x)q(x)}e^{i\gamma_x}\\
    &=&\sum_{x\in S}\cos\gamma_x\sqrt{p(x)q(x)} +
    i\sum_{x\in S}\sin\gamma_x\sqrt{p(x)q(x)},
    \label{EqProjection}
\end{eqnarray}
and finally
\begin{eqnarray}
\nonumber
 |\braket{\xi}{\psi}|^2 &=&
    \left(\sum_{x\in S}\cos\gamma_x\sqrt{p(x)q(x)} \right)^2 +
    \left(
    \sum_{x\in
    S}\sin\gamma_x\sqrt{p(x)q(x)}\right)^2
    \\
\nonumber
    &=&
    \sum_{x,y\in S}\cos(\gamma_x-\gamma_y)\sqrt{p(x)p(y)q(x)q(y)},\\
    &=&
    \sum_{x\in S}p(x)q(x)
    +
    \sum_{x,y\ne x\in
    S}\cos(\gamma_x-\gamma_y)\sqrt{p(x)p(y)q(x)q(y)},
    \label{EqTransitionProb}
\end{eqnarray}
where we used
$\cos(\gamma_x-\gamma_y)=\cos\gamma_x\cos\gamma_y+\sin\gamma_x\sin\gamma_y$
and note that all mixed terms ($\pm
i\cos\gamma_x\sin\gamma_y \ldots$) disappear.

\subsubsection{Length, Euclidian distance of the principal system}
It is useful to compute the Euclidian distance of the principal
system as well in order to compare it with the metric $D$. This is
given by the length of the difference
$\left\|\ket{\psi}-\ket{\xi}\right\|$:
\begin{eqnarray}
\nonumber
  \left\|\ket{\psi}-\ket{\xi}\right\|^2 &=&
  \sum_{x\in S}
  (\alpha_{\varphi_{x}}-\beta_{\chi_{x}})^*(\alpha_{\varphi_{x}}-\beta_{\chi_{x}})\\
\nonumber
   &=& \sum_{x\in S}
   p(x)+q(x) - \sqrt{p(x)q(x)}
   (e^{i\gamma_x}+e^{-i\gamma_x})\\
\nonumber
   &=& \sum_{x\in S} p(x)+q(x) - 2\sqrt{p(x)q(x)} \cos\gamma_x
\end{eqnarray}
with $\gamma_x =\varphi_{x}-\chi_{x}$. Simplifying by using the
normalization condition \ref{EqNormalisationCondition} and Equation
\ref{EqProjection}
we get:
\begin{eqnarray}
  \left\|\ket{\psi}-\ket{\xi}\right\|^2 &=&
  2 - 2 \sum_{x\in S}\cos\gamma_x\sqrt{p(x)q(x)}
   =2 \;\left(1 - \Re\braket{\xi}{\psi}\right).
\label{EqComplexEuklid}
\end{eqnarray}

\subsubsection{Relation between Projection, no-name metric and Euclidian distance of the principal system}
Let $\textbf{P}=\ket{\psi}\bra{\psi}=\textbf{U}^\dag\textbf{DU}$ be
the spectral decomposition of $\textbf{P}$, where $\textbf{D}$ is a
diagonal matrix with non-negative diagonal entries, and $\textbf{U}$
is unitary. Clearly, $\bra{\psi}\textbf{P}\ket{\psi}=1$ and
therefore, $\textbf{U}\ket{\psi} = \ket{k}$ is a basis vector in
some appropriate basis system and $\textbf{D}=\ket{k}\bra{k}$.
Define $\ket{\xi'}= \textbf{U}\ket{\xi}=\sum_{j}\alpha_j'\ket{j}$.
Then, with $\alpha_j' = a_j+ib_j$, $a,b\in \mathbb{R}$,
\begin{equation*}
    \bra{\xi}\textbf{P}\ket{\xi} =
    \bra{\xi}\textbf{U}^\dag\textbf{DU}\ket{\xi} =
    \braket{\xi'}{k}\braket{k}{\xi'}=
    \alpha_{k}'^*\alpha_{k}' = a_k^2+b_k^2,
\end{equation*}
or in terms of the no-name metric we get:
\begin{equation*}
    D\left(\ket{\psi},\ket{\xi}\right)^2 =
1 - a_k^2-b_k^2.
\end{equation*}
The Euclidian distance computes to (see \ref{EqComplexEuklid}):
\begin{eqnarray}
\nonumber
  \left\|\ket{\psi}-\ket{\xi}\right\|^2 &=&
  \left\|\textbf{U}\ket{\psi}-
  \textbf{U}\ket{\xi}\right\|^2
  = \left\|\ket{k}-\ket{\xi'}\right\|^2\\
\nonumber
   &=& 2 \;\left(1 - \Re\braket{\xi'}{k}\right) = 2 \;
   (1 - \Re\alpha_{k}')=2 \;(1 -a_k).
\end{eqnarray}

If we assume a real nonnegative principal system this might give
some insight into distance functions used in the literature
(\cite{Aher97,puzicha:iccv99,Com2003}). In this case as $b=0, 0\le
a\le 1$ and $1-a \le 1-a^2$ we have (see Fig.~\ref{FigExample4}):
$D\left(\ket{\psi},\ket{\xi}\right)\le
\left\|\ket{\psi}-\ket{\xi}\right\| \le
\sqrt{2}D\left(\ket{\psi},\ket{\xi}\right)$.

\subsection{Dropping the phase}\label{PhaseDropping}
\begin{figure}[ht]
    \includegraphics[width=0.8\textwidth]{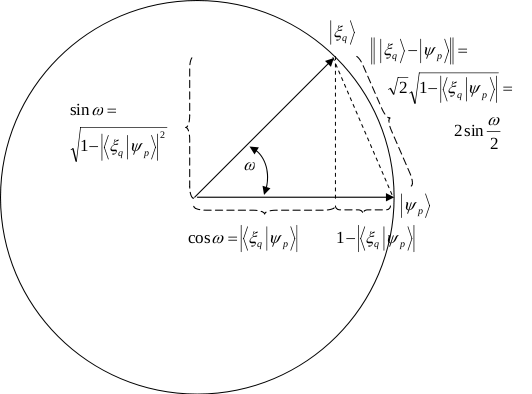}
   \caption{Geometric derivation of several distance measures in $\mathbb{R}^{n}$. The
   intersection of the plane spanned by $\ket{\psi_{{p}}}$ and $\ket{\xi_{{q}}}$
   with the unit hypersphere of $\mathbb{R}^{n}$ generates a unit
   circle.
  }\label{FigExample4}
\end{figure}

The relation between the Euclidian distance of the principal system
and the no-name metric becomes slightly easier (but misleading in
general) when we consider no phase information, i.e. when $p$ and
$q$ are represented by their special nonnegative and real choice
$\ket{\psi_{{p}}}$ and $\ket{\xi_{{q}}}$. In this case the complex
Hilbert space $\mathbb{C}^n$ is reduced to a real Hilbert space of
the same dimension and we may drop the modulus as all probabilities
are mapped onto the positive square root. We can visualize the
relation of the two measures as can be seen in Figure
\ref{FigExample4}.

The inner product $\braket{\xi_{{q}}}{\psi_{{p}}}$ gives the  cosine
of the angle between $\ket{\psi_{{p}}}$ and $\ket{\xi_{{q}}}$ and
equals the length of the projection of $\ket{\xi_{{q}}}$ onto
$\ket{\psi_{{p}}}$. Likewise
$\sqrt{1-\braket{\xi_{{q}}}{\psi_{{p}}}^2}$ results into the sine of
the angle and equals the length of the projection of
$\ket{\xi_{{q}}}$ onto the orthogonal complement of
$\ket{\psi_{{p}}}$ in the plane spanned by the two state vectors.
The length of the difference of the two vectors in terms of the
inner product is given by (Eq. \ref{EqComplexEuklid})
\begin{equation}\label{EqEuclidDistance}
    \|\ket{\psi_{{p}}}-\ket{\xi_{{q}}}\| =
    \sqrt{2}\sqrt{1-\braket{\xi_{{q}}}{\psi_{{p}}}}.
\end{equation}
This relates to a measure between density distributions frequently
used in the literature that is based on the so called Bhattacharyya
coefficients:
\begin{equation}\label{EqFidelity}
    d(p(X), q(X)) \equiv d(p,q) =
    \sqrt{1-\sum_{x\in S}\sqrt{p(x)q(x)}} =
    \sqrt{1-\braket{\xi_{{q}}}{\psi_{{p}}}},
\end{equation}
which up to a constant factor is equal to (\ref{EqEuclidDistance}).
The sum over the Bhattacharyya coefficients, $f(p(X),
q(X))=\sum_{x\in S}\sqrt{p(x)q(x)}$, is usually called the
\textsl{fidelity} of the probability distribution $p$ and $q$. As an
example \ref{EqFidelity} is used in \cite{Com2003} to compare
density functions in the context of object tracking.

Let $\omega$ be the angle between $\ket{\psi_{{p}}}$ and
$\ket{\xi_{{q}}}$. Substituting
$\cos\omega=\braket{\xi_{{q}}}{\psi_{{p}}}$ and observing that
$\sin\frac{\omega}{2} = \frac{1}{\sqrt{2}}\sqrt{1-\cos\omega}$, we
get:
\begin{eqnarray*}
  D(p(X), q(X)) &=& \sin\omega \\
  \|\ket{\psi_{{p}}}-\ket{\xi_{{q}}}\| &=& \sqrt{2} d(p, q) = 2\sin\frac{\omega}{2}.
\end{eqnarray*}

For the rest of the paper we will limit our discussion onto two
dimensional systems as they allow to be visualized
are fairly powerful instruments for concrete
applications.

%
%
%

\section{Visualization of $D$, the two-dimensional case}
\label{ChapQBIT}

It is worthwhile to study the behavior of the transition probability
and the no-name metric more closely. For the two-dimensional Hilbert
space $\mathbb{C}^2$ and pure states $\ket{\psi},\ket{\xi}\in
\mathbb{C}^2$ we can visualize Equation \ref{EqTransitionProb} and
consequently $D(\ket{\psi},\ket{\xi})$.

\begin{figure}[ht]
    \includegraphics[width=0.45\textwidth]{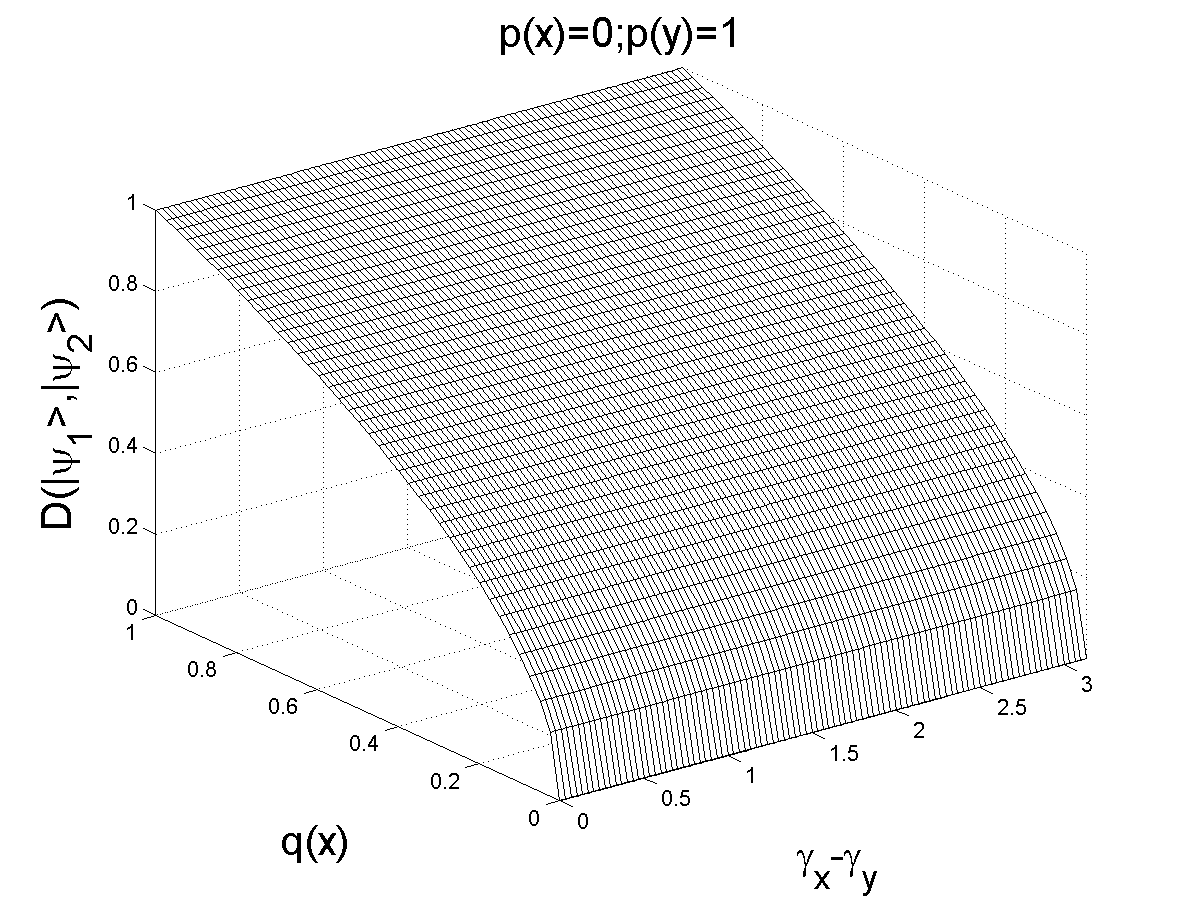}\hfill
    \includegraphics[width=0.45\textwidth]{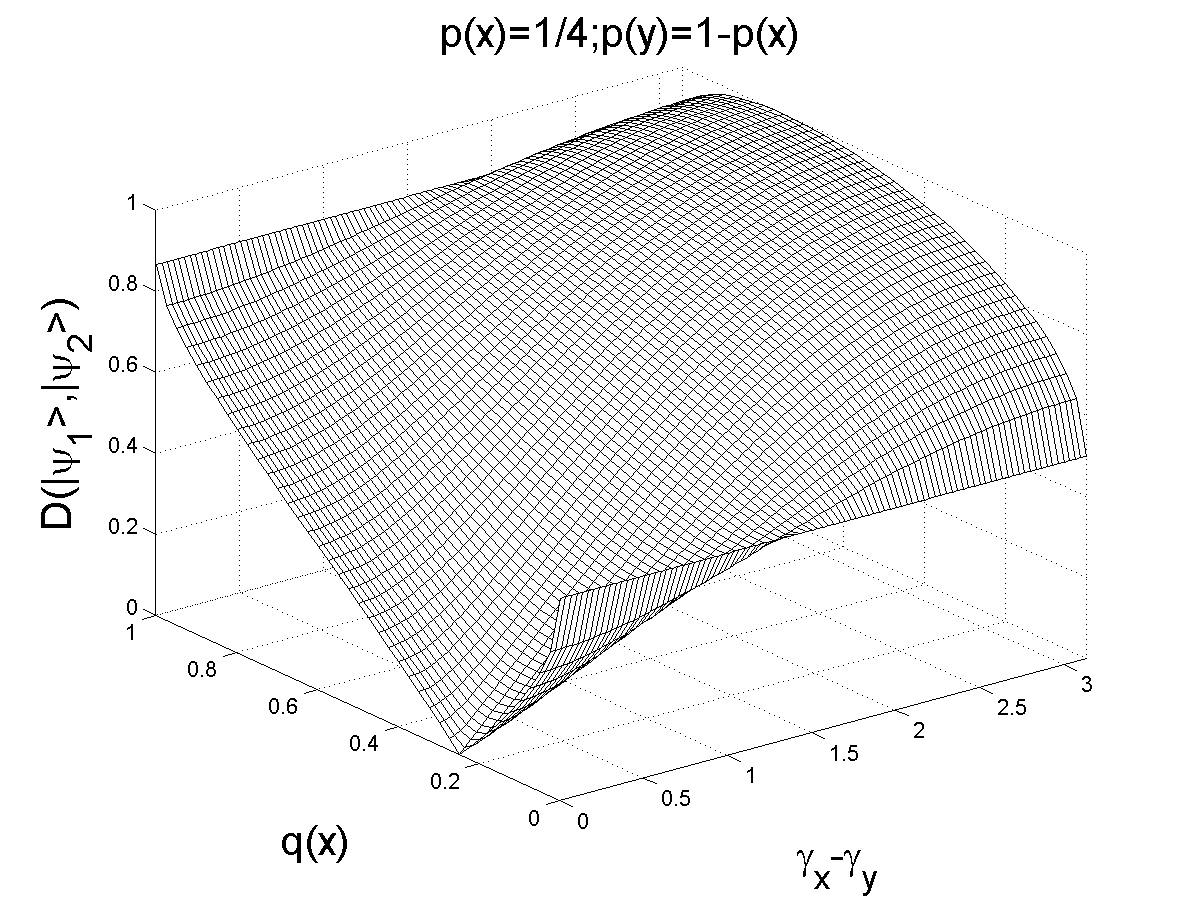}\\
    \includegraphics[width=0.45\textwidth]{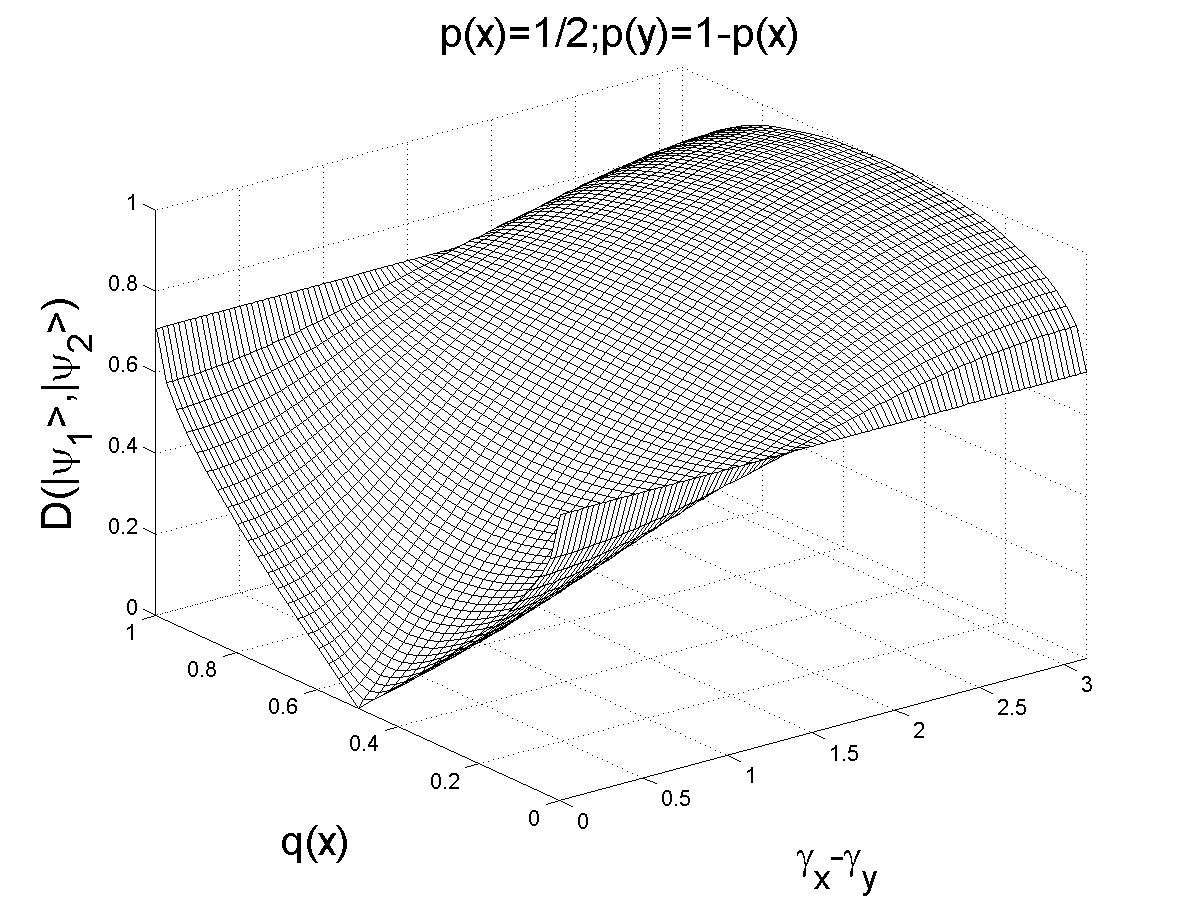}\hfill
    \includegraphics[width=0.45\textwidth]{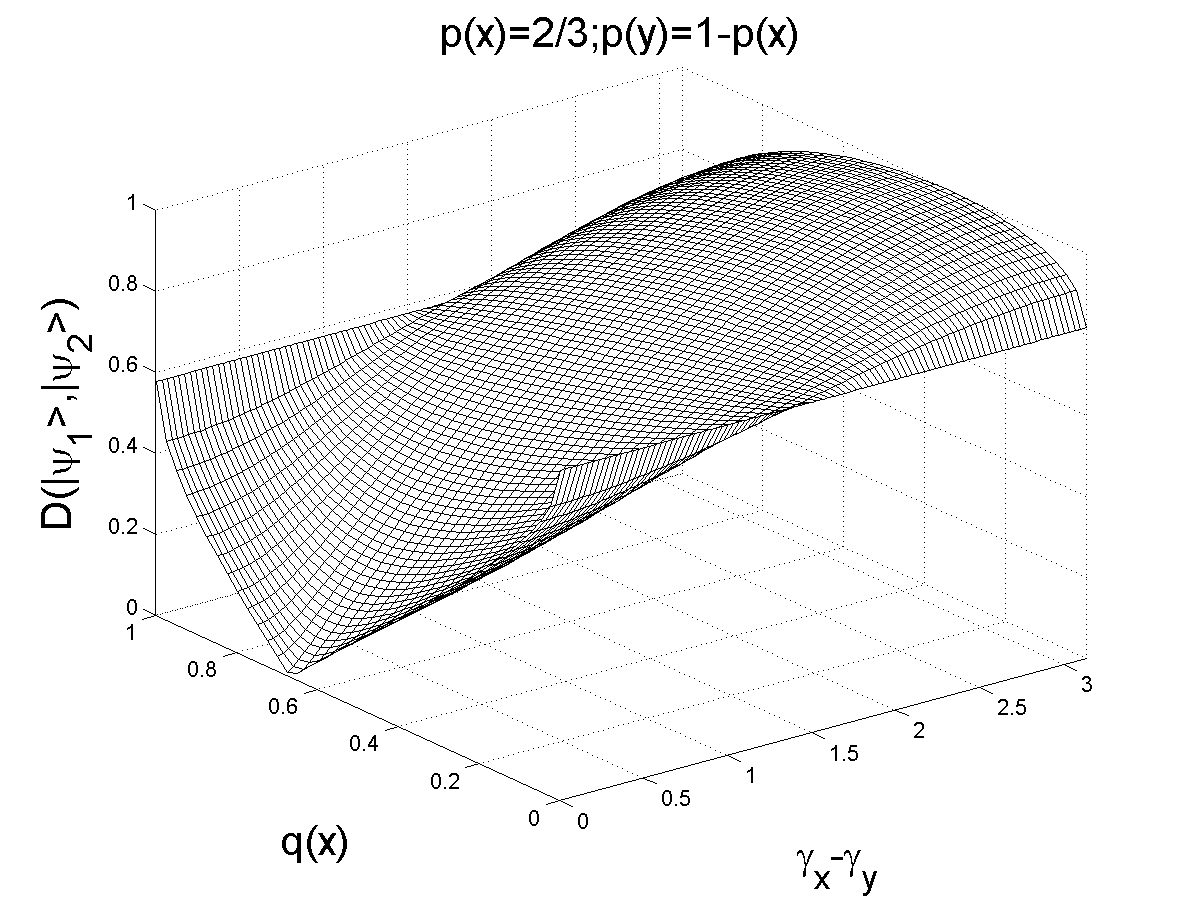}\\
   \caption{Visualization of $D\left(\ket{\psi},\ket{\xi}\right)$ for
   $X = \{x,y\}$ and different values of $p$.
   $q$ ranges over all density distributions and all phase differences between $0$
   and $\pi$. The range from $-\pi$ to $0$ is symmetric.
  }\label{FigExample1}
\end{figure}

\begin{figure}[ht]
    \includegraphics[width=0.45\textwidth]{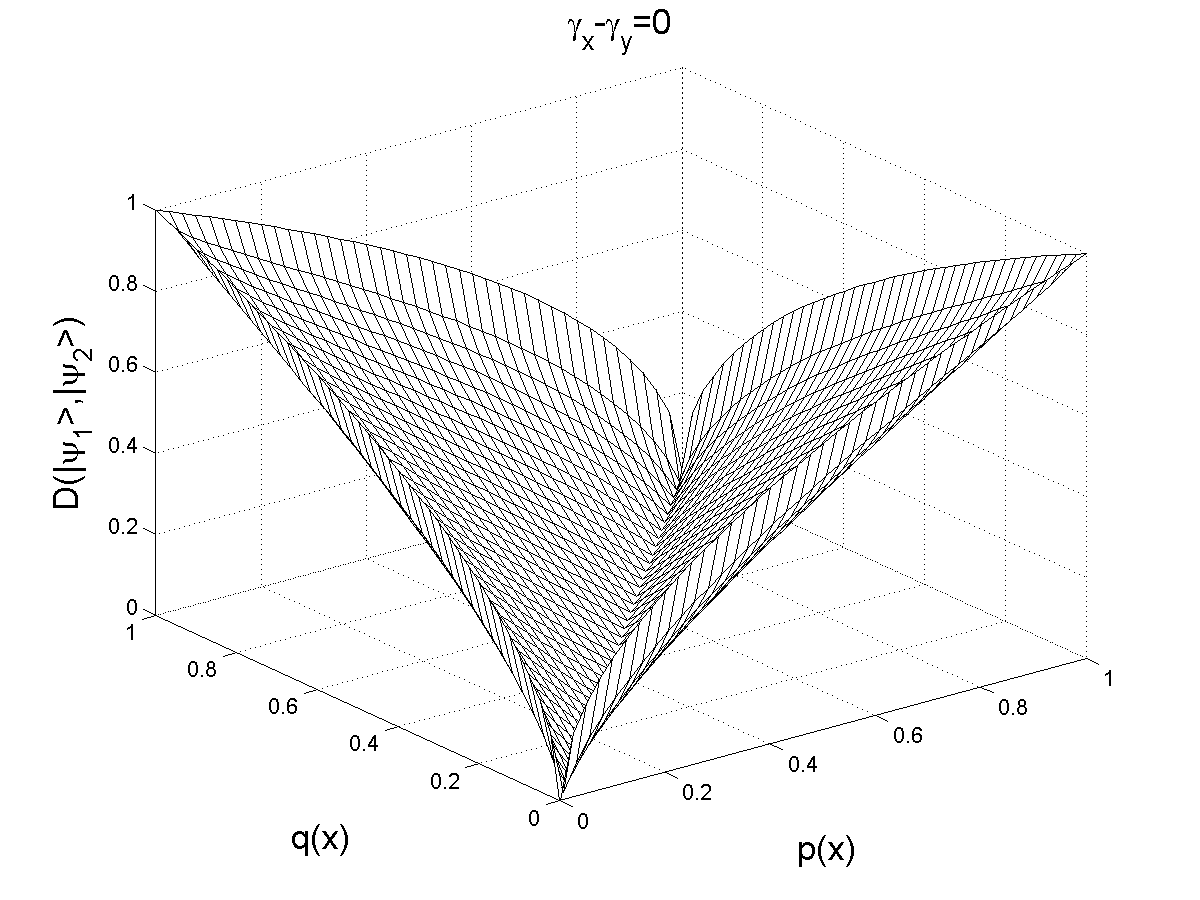}\hfill
    \includegraphics[width=0.45\textwidth]{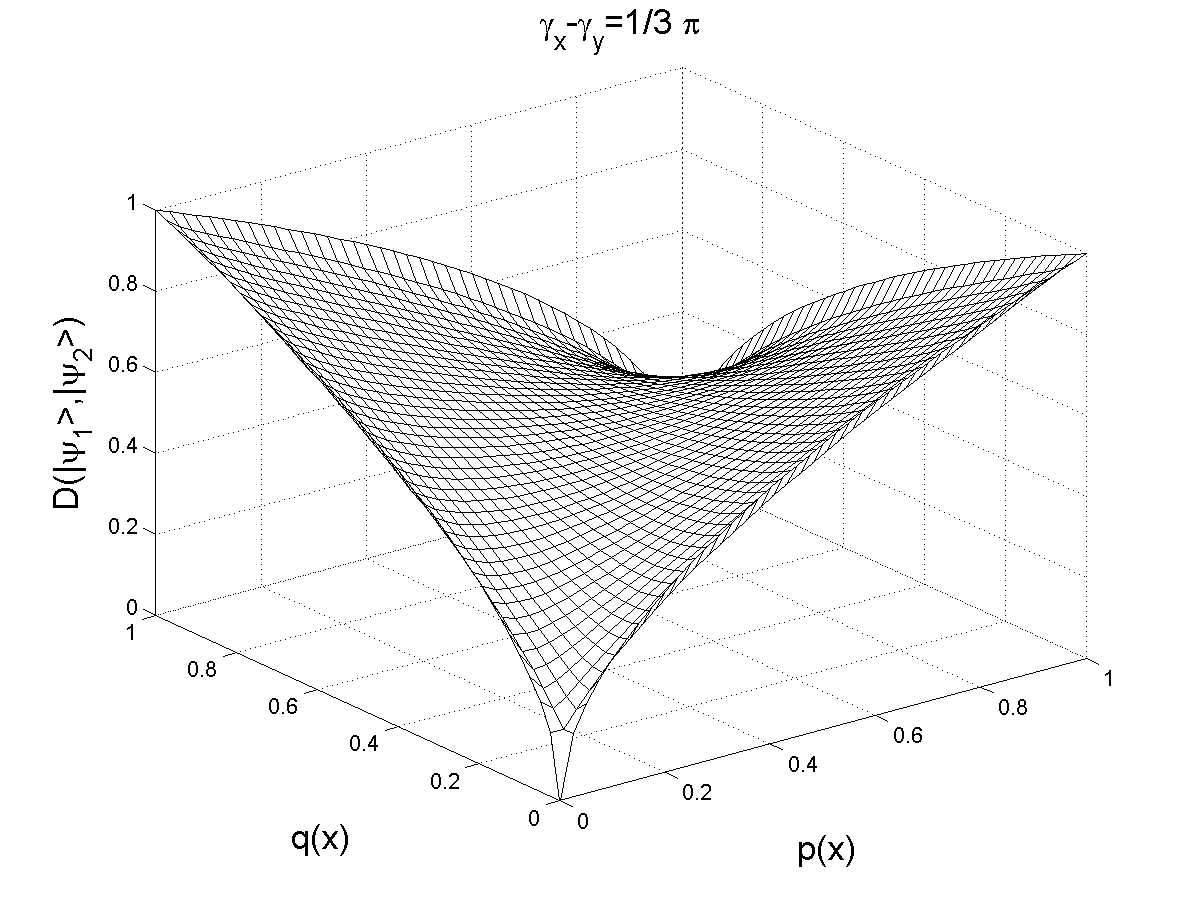}\\
    \includegraphics[width=0.45\textwidth]{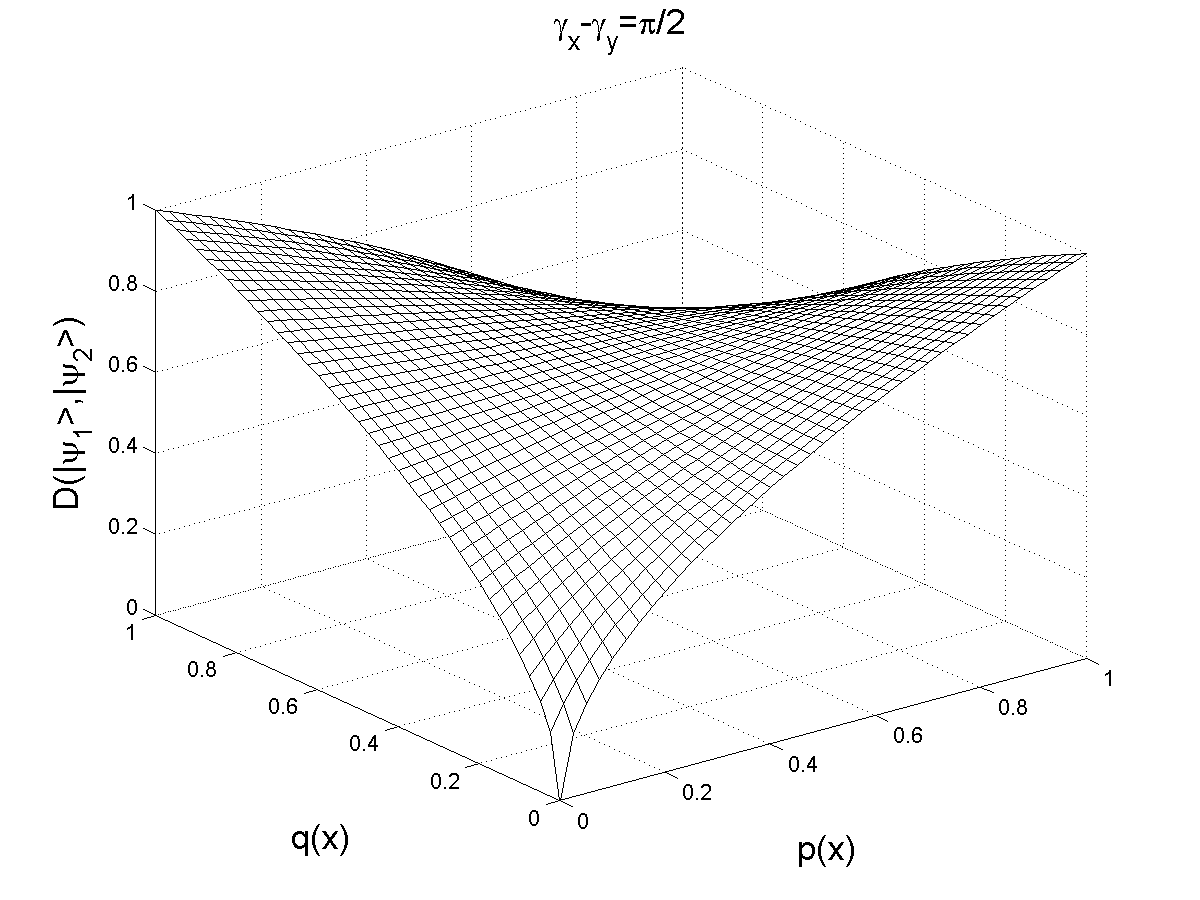}\hfill
    \includegraphics[width=0.45\textwidth]{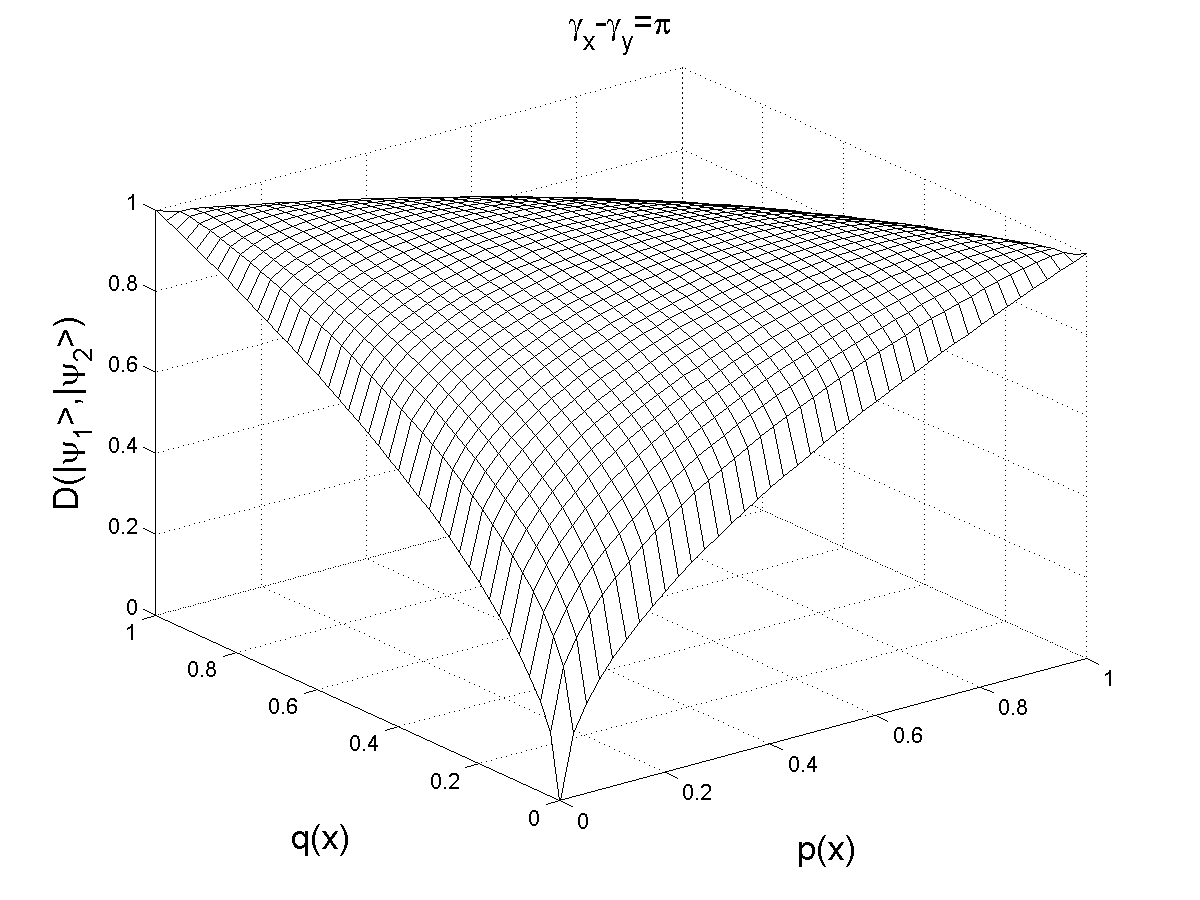}
   \caption{Visualization of $D\left(\ket{\psi},\ket{\xi}\right)$ for $X = \{x,y\}$ and
   a fixed phase difference. $p$ and $q$ range over all possible density distributions.
  }\label{FigExample2}
\end{figure}
\begin{figure}[ht]
    \includegraphics[width=0.45\textwidth]{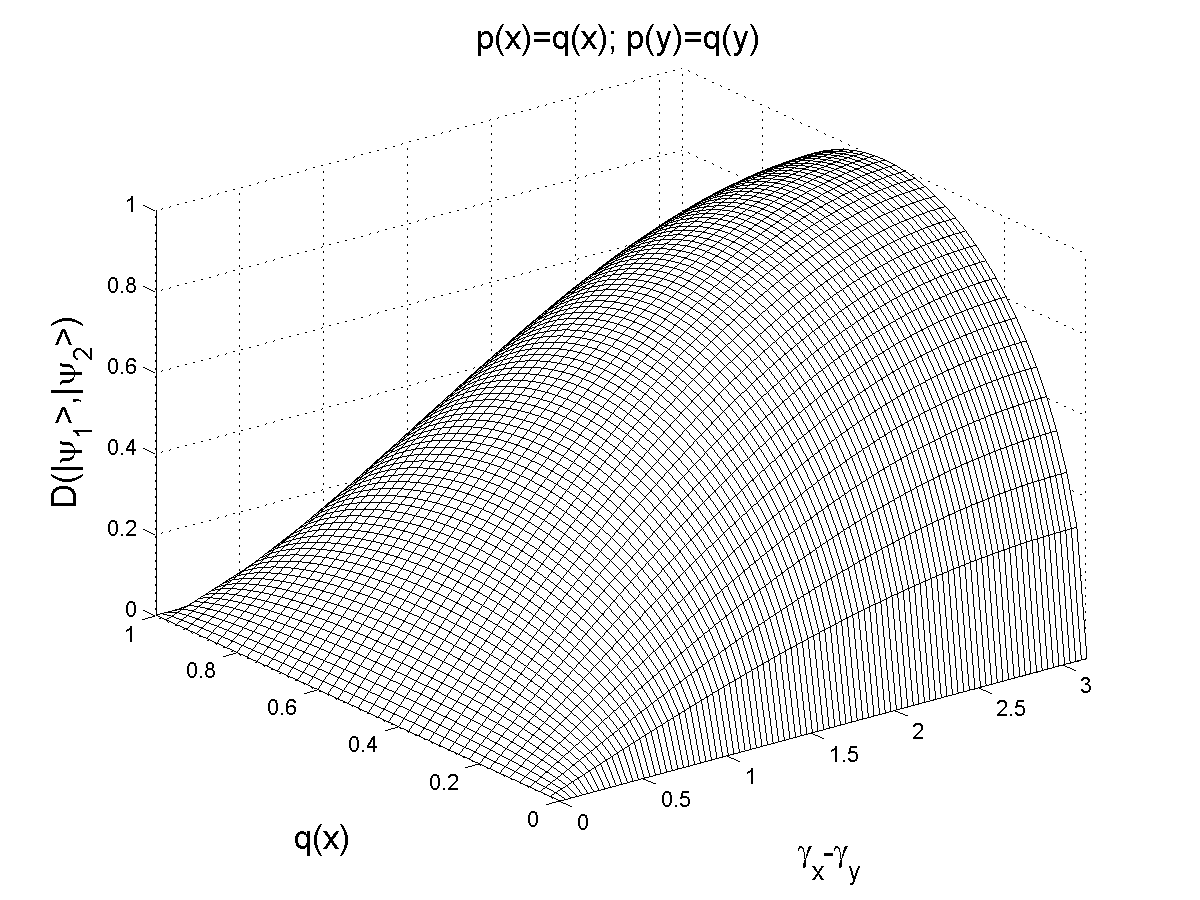}
   \caption{Visualization of $D\left(\ket{\psi},\ket{\xi}\right)$ for $X = \{x,y\}$ and $p=q$.
   The phase difference $\gamma_x-\gamma_y$ ranges over all phase differences between $0$
   and $\pi$. The range from $-\pi$ to $0$ is symmetric.
  }\label{FigExample3}
\end{figure}


Consider density distributions $p, q$ over the random variable
$X = \{x,y\}$. In this case the encoding of $p$ and $q$ as quantum states results in
qubits. We want to analyze the impact of the phase factor on the distance of the qubits.

Using Equation \ref{EqTransitionProb} Equation (\ref{EqFubini}) then takes the form
\begin{equation*}
  D\left(\ket{\psi},\ket{\xi}\right)^2 =
  1 - p(x)q(x)-p(y)q(y)-
  2\cos(\gamma_{x}-\gamma_{y})\sqrt{p(x)p(y)q(x)q(y)}.
\end{equation*}
Alternatively, we may express $D\left(\ket{\psi},\ket{\xi}\right)$
in terms of $p(x)=|\alpha_x|^2$, $p(y)=|\alpha_y|^2$,
$q(x)=|\beta_x|^2$ and $q(y)=|\beta_y|^2$:
\begin{equation*}
  D\left(\ket{\psi},\ket{\xi}\right)^2
  =
  1 - |\alpha_x|^2|\beta_x|^2-|\alpha_y|^2|\beta_y|^2-
  2\cos(\gamma_{x}-\gamma_{y})|\alpha_x||\alpha_y||\beta_x||\beta_y|.
\end{equation*}

\begin{enumerate}
\item Let $p$ be fixed and $q$ vary over all possible
distributions. As $p$ is fixed the phase difference
$\gamma_{x}-\gamma_{y}$ only depends on $q$ and we may consider all
possible phase differences. A visualization of
$D\left(\ket{\psi},\ket{\xi}\right)$ for several choices of $p$ is
given in Figure \ref{FigExample1}.

\begin{enumerate}
    \item Clearly, $D\left(\ket{\psi},\ket{\xi}\right) = 0 \Leftrightarrow \alpha_x=\beta_x$,
$\alpha_y=\beta_y$ as then $\gamma_x-\gamma_y=0$ and
\begin{equation*}
    D\left(\ket{\psi},\ket{\xi}\right)^2 =
    1-(|\alpha_x|^2+|\alpha_y|^2)^2=1-(p(x)+p(y))^2=0.
\end{equation*}

    \item Conversely,
    $D\left(\ket{\psi},\ket{\xi}\right) = 1 \Leftrightarrow
\alpha_x=|\beta_y|e^{i(\chi_{x}\pm \frac{\pi}{2}+\eta)}$,\\
$\alpha_y=|\beta_x| e^{i(\varphi_{q,y}\mp \frac{\pi}{2}+\eta)}$,
$\eta\in\mathbb{R}$, which gives
$\gamma_x-\gamma_y=\pm\pi$, resulting in
\begin{equation*}
D\left(\ket{\psi},\ket{\xi}\right)^2=
1-|\beta_y|^2|\beta_x|^2-|\beta_x|^2|\beta_y|^2+2|\beta_x|^2|\beta_y|^2=1.
\end{equation*}
\end{enumerate}
\item Now let $\gamma_{x}-\gamma_{y}$ be fixed and $p, q$ vary
over all possible distributions. A visualization of
$D\left(\ket{\psi},\ket{\xi}\right)$ for several choices of
$\gamma_{x}-\gamma_{y}$ is given in Figure \ref{FigExample2}.
\begin{enumerate}
    \item $\gamma_{x}-\gamma_{y}=0$:
    \begin{eqnarray*}
        D\left(\ket{\psi},\ket{\xi}\right)^2 &=& 1- p(x)q(x) - p(y)q(y) - 2\sqrt{p(x)p(y)q(x)q(y)}\\
        &=& 1 - \left(\sqrt{p(x)q(x)}+\sqrt{p(y)q(y)}\right)^2\\
        D\left(\ket{\psi},\ket{\xi}\right) &=& 0 \Leftrightarrow p(x)=q(x) \quad\mathrm{and} \\
        D\left(\ket{\psi},\ket{\xi}\right) &=& 1 \Leftrightarrow p(x)=q(y)=0 \vee p(x)=q(y)=1.
    \end{eqnarray*}
    When $p(x)=q(x)$ for all $x\in X$ we have a perfect correlation.
    \item $\gamma_{x}-\gamma_{y}=\pm\frac{\pi}{3}$:
    \begin{eqnarray*}
        D\left(\ket{\psi},\ket{\xi}\right)^2 &=& 1-p(x)q(x)-p(y)q(y)-\sqrt{p(x)p(y)q(x)q(y)}\\
        D\left(\ket{\psi},\ket{\xi}\right) &=& 0 \Leftrightarrow p(x)=q(x)=0 \vee p(x)=q(x)=1 \quad\mathrm{and} \\
        D\left(\ket{\psi},\ket{\xi}\right) &=& 1 \Leftrightarrow p(x)=q(y)=0 \vee p(x)=q(y)=1.
    \end{eqnarray*}
    \item $\gamma_{x}-\gamma_{y}=\pm\frac{\pi}{2}$:
    \begin{eqnarray*}
        D\left(\ket{\psi},\ket{\xi}\right)^2 &=& 1-p(x)q(x)-p(y)q(y)\\
        D\left(\ket{\psi},\ket{\xi}\right) &=& 0 \Leftrightarrow p(x)=q(x)=0 \vee p(x)=q(x)=1 \quad\mathrm{and} \\
        D\left(\ket{\psi},\ket{\xi}\right) &=& 1 \Leftrightarrow p(x)=q(y)=0 \vee p(x)=q(y)=1.
    \end{eqnarray*}

     \item $\gamma_{x}-\gamma_{y}=\pm\pi$:
    \begin{eqnarray*}
        D\left(\ket{\psi},\ket{\xi}\right)^2 &=& 1- p(x)q(x) - p(y)q(y) + 2\sqrt{p(x)p(y)q(x)q(y)}\\
        &=& 1 -\left(\sqrt{p(x)q(x)}- \sqrt{p(y)q(y)}\right)^2\\
        D\left(\ket{\psi},\ket{\xi}\right) &=& 0 \Leftrightarrow p(x)=q(x)=0 \vee p(x)=q(x)=1 \quad\mathrm{and} \\
        D\left(\ket{\psi},\ket{\xi}\right) &=& 1 \Leftrightarrow p(x)=1-q(x).
    \end{eqnarray*}
    When $p(x)=1-q(x)$ for all $x\in X$ we have a perfect anti correlation.
\end{enumerate}

\item Figure \ref{FigExample3} shows $D\left(\ket{\psi},\ket{\xi}\right)$ if $p=q$ and $q$ is
ranging over all phase differences between $0$ and $\pi$.
\end{enumerate}

\subsection{Visualization of qubits, the Bloch sphere}
A very useful visualization technique that is limited to two
dimensional systems is given by the so called Bloch sphere. 
By ignoring a physically irrelevant overall phase factor, the general
state of a qubit can be written as
\begin{equation}
  \kpsi=\cos\frac\theta2 \ket{0}+e^{i\varphi}\sin\frac\theta2 \ket{1}
  \mwith \varphi\in[0,2\pi),\,\theta\in[0,\pi]. \label{Blochvector}
\end{equation}

If we interpret $\theta$ and $\varphi$ as spherical coordinates
\begin{equation}
  \hat{r}=(\cos\varphi\sin\theta,\sin\varphi\sin\theta,\cos\theta)=(x,y,z),
  \label{EqBlochPolar}
\end{equation}
every qubit state has a unique representation as a point on the
three-dimensional unit sphere, also known as {\it Bloch sphere} (see
Fig. \ref{FigBlochSphere}).

The unit-vector $\hat{r}=\hat{r}_\psi$ is called {\it Bloch vector}
of $\kpsi$. Bloch vectors have the property that
\begin{equation}
  \hat{r}_\phi = -\hat{r}_\chi \iff \braket{\phi}{\chi}=0.
\end{equation}

For density operators of a two-dimensional system a similar
generalization exists. By again ignoring an overall phase, any
single qubit density operator can be written as (see \cite{IQC05})
\begin{equation}
  \rho=\frac{1}{2}\matr{cc}{1+z&x-iy\\x+iy&1-z}.
\end{equation}

For mixed states, the Bloch vector $r_\rho=(x,y,z)$ lies inside the
Bloch sphere ($\|r_\rho\|<1$) and for a classical unbiased coin flip
state, i.e. $p(\ket{0})=p(\ket{1})=1/2$ we get
\begin{equation}
 \rho_c=\matr{cc}{\frac{1}2&0\\0&\frac{1}2}
 \quad \Rightarrow r_{\rho_c}=(0,0,0).
\end{equation}

Unfortunately, there is no easy generalization of the Bloch sphere
for higher dimensional quantum systems.

%
%
%

\section{A New Approach to Signal Segmentation using Conjugate Information Variables}\label{ChapApproach}

The segmentation of signals is one of the fundamental
problems in the area of signal and image processing. This is
especially true as many higher-level signal analysis algorithms rely
heavily on the result of a low-level segmentation process. The term
segmentation is mostly not well defined as it can refer to finding
some objects in an image, e.g. a face, skin, coin, etc., or
detecting lower-level features like edges, constant signal areas,
etc.

\begin{figure}[ht]
    \includegraphics[width=0.45\textwidth]{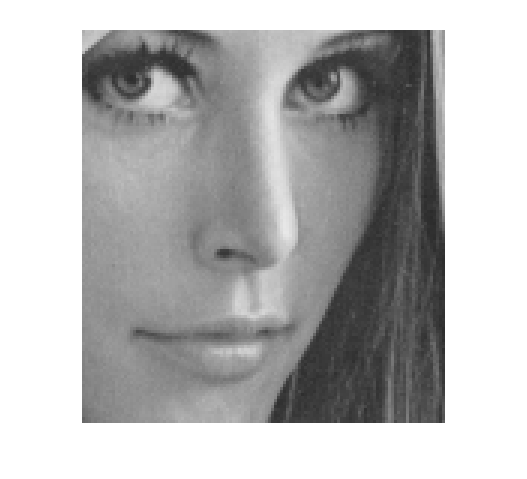}\hfill
    \includegraphics[width=0.45\textwidth]{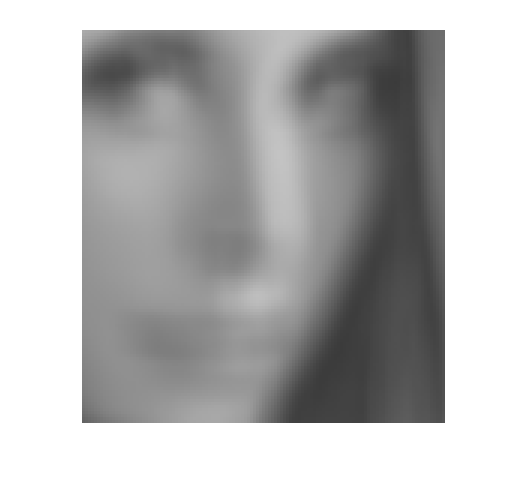} \\
    \includegraphics[width=0.45\textwidth]{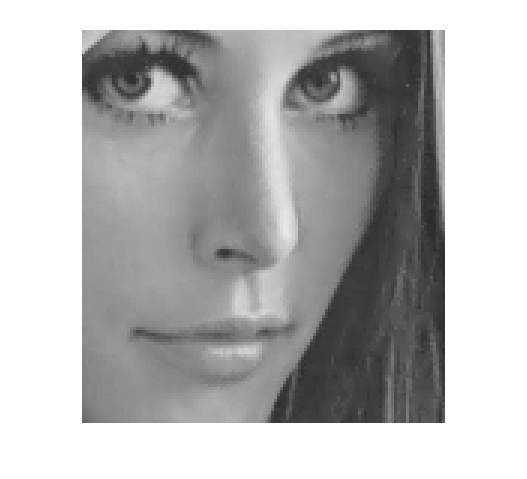}\hfill
    \includegraphics[width=0.55\textwidth]{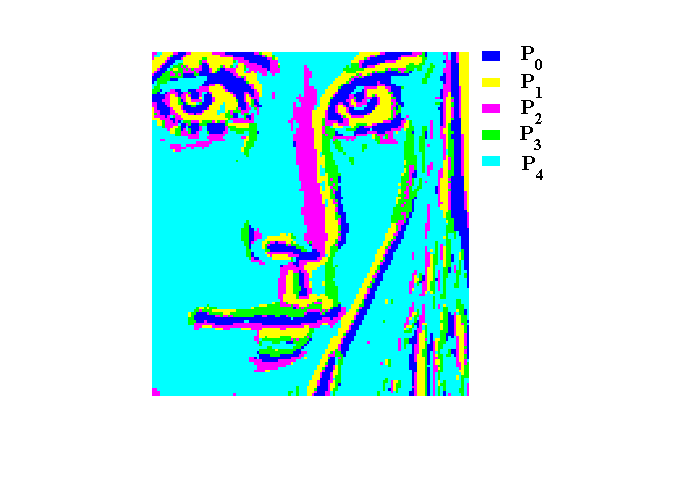}
   \caption{An application of the segmentation technique. The original image on the top left
    was smoothed on the constant part $P_4$ of the image function as given
    in the bottom right picture. The result of smoothing adaptively with an average filter ($15\times 15$)
    is shown in the lower left picture ($P_0,\dots,P_3$ are not filtered).
    The top right shows the filter result using the same filter kernel
    independent of the image content.
  }\label{FigLenna}
\end{figure}

In this chapter we suggest a new approach for a specific
segmentation problem, namely the segmentation of signals into
locally constant, rising or falling parts, and minima or maxima. For
simplicity we assume one-dimensional signals but remark that
the approach is not limited to this and can be extended to two- or 
higher-dimensional problems (see Fig. \ref{FigLenna}).

Let the signal be given by $f(t)$ and let it be at least twice
differentiable, that is $f^{'}(t)$ and $f^{''}(t)$ exist. 
In our approach the deriviates will be used as as conjugate variables in order
to classify local parts of functions.

We need the range of the derivatives
to be within certain limits to normalize them properly. This can
sometimes cause problems as we have to think about meaningful
limits in a given application but usually the
limits are obvious for real life signals, such as an 
image function where the pixel's range from zero to one, etc.
The limits of the derivatives will be used as soft thresholds that control each other.

Zero crossings of the first derivative indicate extrema of a function and the sign of the 
second dervative destincts between a maximum and a minimum. 
The function is constant or has a saddle point if both derivatives are zero.
In the presence of noise zero crossings of the first derivative becomes unstable 
whereas the sign of the second derivative remains robust unless there is a saddle point.
We will expoit this fact and relate it the role amplitude and phase play for qubits. 

Let
\begin{eqnarray*}
    p^\pm(t) &=& \sqrt{\frac{c_{2} \pm f^{''}(t) }{2 c_{2}}},
     \\
    \alpha^{+}(t) &=& p^{+}(t) e^{i \pi f^{'}(t)/c_1}, \\
    \alpha^{-}(t) &=& p^{-}(t), \\
    &&\textrm{with $|f^{'}(t)| \le c_1, |f^{''}(t)| \le c_2$ for all
    $t$,
    }
\end{eqnarray*}
and define quantum states
\begin{equation}\label{R2Qubit}
    \kpsix{t} = \alpha^{+}(t)\ket{z_+} + \alpha^{-}(t)\ket{z_-}.
\end{equation}

\begin{figure}[ht]
    \includegraphics[width=0.45\textwidth]{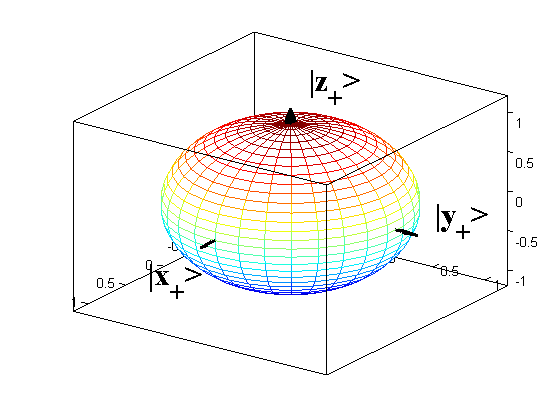}\hfill
    \includegraphics[width=0.45\textwidth]{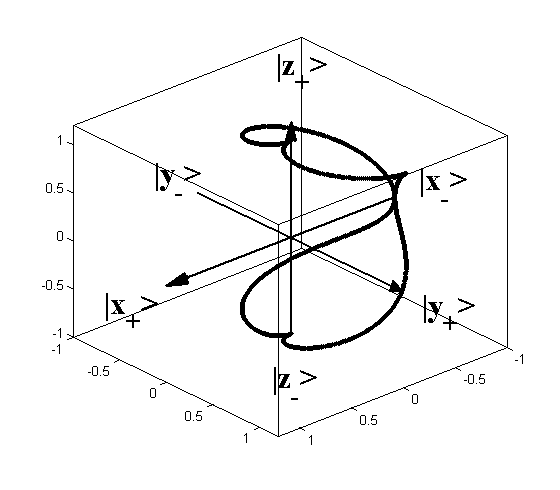}\\
    \includegraphics[width=0.45\textwidth]{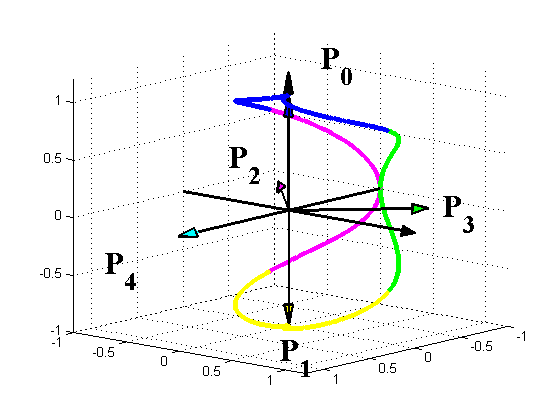}\hfill
    \includegraphics[width=0.45\textwidth]{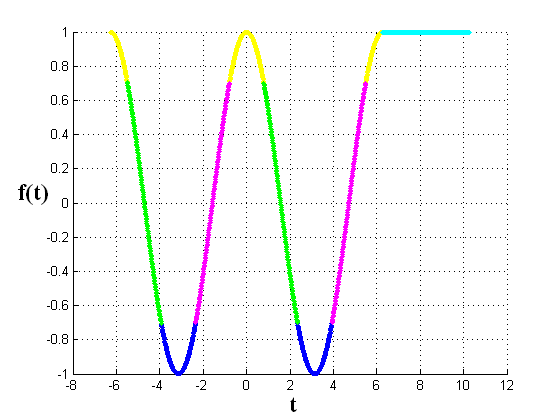}\\
    \includegraphics[width=0.32\textwidth]{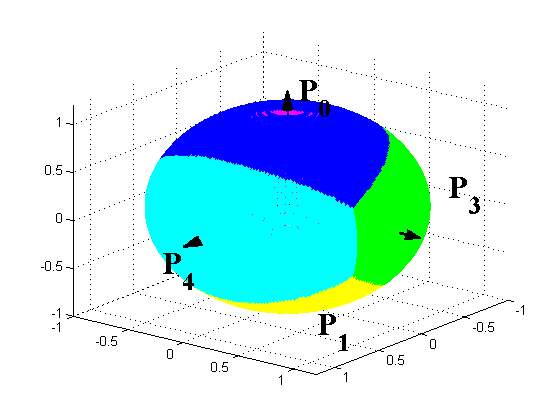}\hfill
    \includegraphics[width=0.32\textwidth]{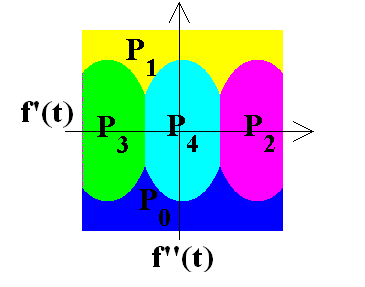}\hfill
    \includegraphics[width=0.32\textwidth]{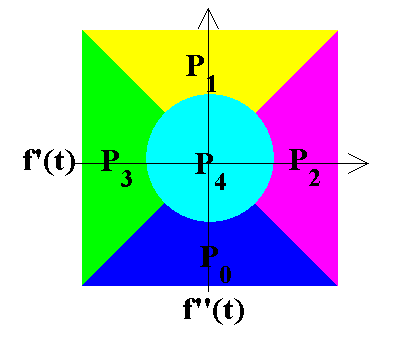}
   \caption{Top left: Bloch sphere representation of all qubit states.
   Top right: Bloch sphere representation of the qubit states
   $\kpsix{t}$ for $f(t)$ as given in Equation \ref{cosconst}.
   Middle left: Labeling result in the Bloch sphere representation.
   Middle right: Labeling of $f(t)$.
   Bottom left: Decision regions in the Bloch sphere representation,
   $P_2$ is shining through at the north pole but is not visible from this perspective.
   Bottom middle: Decision regions over the joint distribution of $f^{''}(t)$ and
   $f^{'}(t)$. Bottom right: Model based (see Equations \ref{EqModel}) decision regions over the joint distribution of $f^{''}(t)$ and
   $f^{'}(t)$. The limits of the axes are $\pm c_1$ for $f^{'}(t)$
   and $\pm c_2$ for $f^{''}(t)$.  If $c=c_1=c_2$, the radius of the inner circle,
   $P_4$, is given by $c/2$.
  }\label{FigBlochSphere}
\end{figure}

Please note that as $|\alpha^{+}(t)|^2+|\alpha^{-}(t)|^2=1$, for all
$t$, the $\kpsix{t}$ are well-defined qubits. $\ket{z_+}, \ket{z_-}$
describe the positive and negative $z$-axis of the Bloch sphere (see
Fig. \ref{FigBlochSphere}) and $f^{''}(t) = \pm c_2
\Longleftrightarrow \kpsix{t} = \ket{z_\pm}$ independent of
$f^{'}(t)$. The positive and negative $x$ and $y$ axis respectively
are given by
\begin{eqnarray*}
    \ket{x_+}& = \frac{1}{\sqrt{2}}(\ket{z_+}+\ket{z_-}),\quad
    &\ket{x_-} = \frac{1}{\sqrt{2}}(\ket{z_+}-\ket{z_-}),\\
    \ket{y_+} &= \frac{1}{\sqrt{2}}(\ket{z_+}+i\ket{z_-}),\quad
    &\ket{y_-} = \frac{1}{\sqrt{2}}(\ket{z_+}-i\ket{z_-}).
\end{eqnarray*}
$f^{''}(t) = f^{'}(t) = 0 \Longleftrightarrow \kpsix{t}=\ket{x_+}$
whereas $f^{''}(t) = 0 \wedge f^{'}(t) = \pm c_1 \Longleftrightarrow
\kpsix{t}=\ket{x_-}$. $f^{'}(t)$ and $f^{''}(t)$ are now conjugate
variables in the sense that observing the random variables
$X=\{\ket{x_+}\bra{x_+},\ket{x_-}\bra{x_-}\}$ and
$Z=\{\ket{z_+}\bra{z_+},\ket{z_-}\bra{z_-}\}$ the
commutator\footnote{The commutator of two operators $A, B$ is given
by $[ A,B ] = AB-BA$. Two operators commute if and only if their
commutator equals zero, i.e. $AB = BA \Longleftrightarrow [ A, B ] =
0.$} between any two projectors of $X$ and $Z$ becomes maximal:
\begin{equation*}
    A_\pm=\ket{x_\pm}\bra{x_\pm},
    B_\pm=\ket{z_\pm}\bra{z_\pm}\Rightarrow
     \|A_\pm B_\pm - B_\pm A_\pm \| = \frac{1}{2}.
\end{equation*}

Next we define a set of projectors:
\begin{eqnarray*}
  P_0 &=& \ketbra{z_+}{z_+} \\
  P_1 &=& \ketbra{z_-}{z_-} \\
  P_2 &=& \frac{1}{\sqrt{2}}\left(\ketbra{y_-}{y_-}+\ketbra{x_-}{x_-}\right)
       = \frac{1}{\sqrt{2}}\left( P_0-\frac{1}{\sqrt{2}}(1+i)P_1\right)\\
  P_3 &=& \frac{1}{\sqrt{2}}\left(\ketbra{y_+}{y_+}+\ketbra{x_-}{x_-}\right)
       = \frac{1}{\sqrt{2}}\left( P_0-\frac{1}{\sqrt{2}}(1-i)P_1\right)\\
  P_4 &=& \ketbra{x_+}{x_+} = \frac{1}{\sqrt{2}}\left(
  P_0+P_1\right)= \frac{1}{\sqrt{2}}\mathbb{I}.
\end{eqnarray*}
Only $P_0$ and $P_1$ form a random variable as they add up to
the identity and therefore fulfill the completeness condition. 
They indicate minimum or maximum of the function. 
$P_i, i=2,\ldots,4$ would have to be complemented by $\mathbb{I}-P_i$ in
order to do so. The choice of projectors depends on the
problem we want to solve.

For any given $f(t)$ we now can project the corresponding
$\ket{\psi(t)}$ and get a classification result by maximizing over
all projectors
\begin{equation*}
    C(f(t)) = i\in \{0,\ldots,4\}:
    \bra{\psi(t)}P_i\ket{\psi(t)}=\max_{j=0,\ldots,4}\bra{\psi(t)}P_j\ket{\psi(t)}.
\end{equation*}
Alternatively, if $X_i =\{P_i, \mathbb{I}-P_i\} =
\{\ket{\chi_{i,0}}\bra{\chi_{i,0}},\ket{\chi_{i,1}}\bra{\chi_{i,1}}\},
i=0,\ldots,4$, and $\ket{\chi}$ represents the eigenvector of the corresponding projector, 
we can minimize the distance
$D(\ket{\chi_{i,0}},\ket{\psi(t)})$.
The results of the classification for an analytic function
\begin{equation}\label{cosconst}
    f(t) = \left\{\begin{array}{cc}
             \cos(t), & |t|<2\pi \\
             1, & t>2\pi
           \end{array}\right.
\end{equation}
are given in middle row of Fig.~\ref{FigBlochSphere}, whereas the
bottom row (left and middle picture) shows the decision boundaries
over the entire joint distribution of the derivatives as derived by
the described approach. Usually we cannot access an analytic
expression for $f(t)$ but a measurement of either $f(t)$ or
$f^{'}(t)$ and $f^{''}(t)$ is available. If we measure $f(t)$ we have
to derive the derivatives numerically\footnote{Depending on the
application, this might require some low pass filtering of the
measurements to gain stability in the derivatives.}. An example for the 
measurement of $f^{'}(t)$ and $f^{''}(t)$
is given by observing the velocity and acceleration of an object. 
In general these measurement will be taken by independent devices but they are highly 
correlated quantities. 

In either case the measurements will be subject to errors. We may
choose different error models for the respective areas.
In the example we gave in the introduction we
could decide that the acoustic signal is mainly noise during the day
time and the visual signal is not available at night time. Towards
the poles of the Bloch sphere the phase term (in this case
$f^{'}(t)$) becomes less important as it seamlessly transforms
itself into a global phase factor. In equatorial regions it becomes
predominant. A quantitative discussion of this depends on the
concrete problem to be solved, especially on the statistical error
model(s) to be chosen, and shall be left for a future publication.

A standard approach to derive the decision regions would be to
create a model 
of the decision problem
and than minimize some distance function
over the entire joint signal distribution to this model as
shown in the bottom right picture of Fig.~\ref{FigBlochSphere} for
the function given in Equation \ref{cosconst}. To derive this model
we have to notice the (partially) nonlinear dependency of
\begin{eqnarray}
 \nonumber f^{'}(t)= -\sin(t),\quad  f^{''}(t) = -\cos(t),\quad
  f^{'}(t)^2+f^{''}(t)^2 = 1, &&|t|<2\pi,\\
 f^{'}(t)=f^{''}(t)=0, &&t\ge2\pi. \label{EqModel}
\end{eqnarray}

In this model the decision depends on both signals over (almost) the
entire joint distribution range, whereas the model derived from the
conjugate variables cuts off the influence of one signal around the
poles of the Bloch sphere.

\section{Discussion}\label{ChapDis}

The approach of modelling joint distributions of signals as conjugate
variables certainly has its limitations. We remark that its
strengthes go along with its weaknesses. Whenever all signals are
meaningful over the entire joint distribution range other approaches
might be more successful\footnote{In the Hilbert space setting used
in this paper we need a higher dimensional system, e.g. a two qubit
system. Of course, we may add conjugate information to each of those
systems, if it is available to us.}.

On the other hand in situations where we have an expectation that we
can rely on some signals more than others, depending on some given
(or derived) parameters, our approach will be valuable as this
dependency may be modelled directly. It is noteworthy that the role
of the two signals used for the encoding might be quite different.
In the bird example we may classify the two species, lets call them
{\sl Happy} and {\sl Unhappy}, using the acoustic signal alone at
times but even when this signal becomes mixed we still can rely on
its presence and then take the optical signal into account. The
converse is not true as we can not speak about a colour at night.
Therefore, the optical sensor (used as phase information) might give
arbitrary measurement results (including blue or green!) at night
time. This will not influence the classification result as the
measurement of the conjugate acoustic signal is mapped to one of the
polar regions at night time. In lucky circumstances, e.g. when by chance
we have no background noise during the day or the moon provides
enough light at night, the joint observation of chirp and colour
does not change the classification result either, as a {\sl Happy}
bird does not become more {\sl Happy} if we know both features. When
the noise level is slowly increasing (at sunrise or sunset)
seamlessly both signals contribute to the classification and allow a
discrimination of the two classes.

Just to give a flavour of the results to expect, Fig.~\ref{FigLenna}
shows the application of an adaptive filter on image functions that
is based on the segmentation approach. At the top left the
original image (used as the signal $f(t)$) is given which is to be
smoothed to remove noise introduced by the sensor (camera, scanner,
etc.). If we assume uniform noise, we can remove this by averaging
over small parts of the image function. The application of an
average filter independent of the image structure will remove this
noise but unfortunately it will remove parts of the image content at
the same time as can be seen in the top right picture of
Fig.~\ref{FigLenna}. If we classify each point of the image
according to the approach described in Section \ref{ChapApproach},
we can apply the averaging process only to image areas that are
labelled constant, i.e. $P_4$ (Fig.~\ref{FigLenna} bottom right).
This results in an image that is fairly smooth in these constant
image areas but untouched everywhere else (Fig.~\ref{FigLenna}
bottom left). In the other areas we could apply different filter
kernels (and different error models) if we wish to do so.

\section{Conclusion and Future Work}\label{ChapFut}

In this paper we have described a method for modelling joint
distributions of signals as conjugate variables in finite-dimensional 
complex Hilbert spaces. We have derived a distance
function based on the transition probability of quantum states that
is a metric on pure states and satisfies the triangle inequality for
mixed states, and we have related it to some distance functions between
probability distributions. The analysis of other distance measures
on quantum states, especially the so called trace distance which is
closely related to the variational distance of probability
distributions, will be a future task.

We believe that the mathematical concept of quantum information
theory offers a fertile resource in many areas of information
processing.
In this article we mainly focused on
non-decomposed principle systems, but it would be very
interesting to analyze the decomposition of higher-dimensional
systems which remains a future task. We have not touched on the field
of quantum operations, i.e unitary transformations of quantum states
which again offers a rich potential. Additionally generalized
measurement operators (positive operator valued measurements (POVM))
offer a valuable future research direction. The use of Wigner
functions provides another promising research area. Wigner functions
relate the probability distributions of conjugate variables to each
other as they describe this joint (real valued) distribution.
Moreover they relate the spatial to the frequency domain, which may
have further impact.

\bibliographystyle{amsplain}
\providecommand{\bysame}{\leavevmode\hbox to3em{\hrulefill}\thinspace}
\providecommand{\MR}{\relax\ifhmode\unskip\space\fi MR }
\providecommand{\MRhref}[2]{%
  \href{http://www.ams.org/mathscinet-getitem?mr=#1}{#2}
}
\providecommand{\href}[2]{#2}

\end{document}